%% file: template.tex
\documentclass{article}

\usepackage{arxiv}

\usepackage[utf8]{inputenc} 
\usepackage[T1]{fontenc}    
\usepackage{hyperref}       
\usepackage{url}            
\usepackage{booktabs}       
\usepackage{amsfonts}       
\usepackage{nicefrac}       
\usepackage{microtype}      
\usepackage{lipsum}
\usepackage{graphicx}
\usepackage{amsmath,amsfonts,amssymb}
\usepackage{algorithmic}
\usepackage{algorithm}
\usepackage{array}
\usepackage[caption=false,font=normalsize,labelfont=sf,textfont=sf]{subfig}
\usepackage{textcomp}
\usepackage{stfloats}
\usepackage{url}
\usepackage{verbatim}
\usepackage{graphicx}
\usepackage{cite}
\usepackage{wrapfig}

\usepackage{tabularx}
\usepackage{siunitx} 
\usepackage{xspace}
\usepackage[table]{xcolor}
\usepackage{booktabs,multirow,makecell}
\usepackage{listings}
\usepackage{enumitem}
\usepackage[most]{tcolorbox}

\usepackage{xcolor}
\usepackage{soul}

\newtcblisting{promptbox}{
  breakable,
  listing only,
  colback=gray!5,
  colframe=gray!40,
  boxrule=0.4pt,
  enhanced,
  left=4pt, right=4pt, top=3pt, bottom=3pt,
  before skip=4pt,
  after skip=8pt,
  listing options={
    basicstyle=\ttfamily\footnotesize,
    breaklines=true,
    breakatwhitespace=false,
    columns=fullflexible,
    showstringspaces=false
  }
}

\input{math.tex}

\graphicspath{ {./images/} }

\title{PDA: Text-Augmented Defense Framework for Robust Vision-Language Models against Adversarial Image Attacks}

\author{
 Jingning Xu, Haochen Luo, Chen Liu\thanks{Corresponding author} \\
Department of Computer Science \\
City University of Hong Kong \\
  \texttt{ \{jingninxu3-c,chester.hc.luo\}@my.cityu.edu.hk, chen.liu@cityu.edu.hk} \\
}

\begin{document}
\maketitle
\begin{abstract}
Vision–language models (VLMs) are vulnerable to adversarial image perturbations.
Existing works based on adversarial training against task-specific adversarial examples are computationally expensive and often fail to generalize to unseen attack types.
To address these limitations, we introduce \textit{Paraphrase–Decomposition–Aggregation (PDA)}, a training-free defense framework that leverages text augmentation to enhance VLM robustness under diverse adversarial image attacks. 
PDA performs prompt paraphrasing, question decomposition, and consistency aggregation entirely at test time, thus requiring no modification on the underlying models.
To balance robustness and efficiency, we instantiate PDA as invariants that reduce the inference cost while retaining most of its robustness gains.
Experiments on multiple VLM architectures and benchmarks for visual question answering, classification, and captioning show that PDA achieves consistent robustness gains against various adversarial perturbations while maintaining competitive clean accuracy, establishing a generic, strong and practical defense framework for VLMs during inference.

\end{abstract}


\input{sec/1_intro}

\input{sec/2_related_work}
\input{sec/3_methodology}

\input{sec/4_exp}
\input{sec/5_conclusion}

\bibliographystyle{abbrv}
\bibliography{main}


\input{sec/X_suppl}
\end{document}

%% file: math.tex
\usepackage{amsmath,amsfonts,bm,amssymb,bbm}

\def\eqref#1{equation~\ref{#1}}

\def\1{\bm{1}}

\DeclareMathAlphabet{\mathsfit}{\encodingdefault}{\sfdefault}{m}{sl}
\SetMathAlphabet{\mathsfit}{bold}{\encodingdefault}{\sfdefault}{bx}{n}

\def\gB{{\mathcal{B}}}

\def\gL{{\mathcal{L}}}

\def\gS{{\mathcal{S}}}
\def\gT{{\mathcal{T}}}



%% file: sec/1_intro.tex
\section{Introduction}
\label{sec:intro}

Vision Language Models (VLMs) have recently achieved substantial success in both general-purpose multimodal systems~\cite{Liu2023LLaVA,Li2023BLIP2,Chen2024PaLIX,Ye2024mPLUGOwl2} and applications in specific domains, including medical imaging~\cite{Bannur2023BioViLT,Wan2023MedUniC,Lin2023PMC_CLIP}, autonomous driving~\cite{Sima2024DriveLM,Qian2024nuScenesQA}, and robotics control~\cite{Jiang2023VIMA,Brohan2023RT2}.
Despite these advances, VLMs are shown susceptible to small, human-imperceptible image-space perturbations that induce large prediction shifts, posing outsized deployment risks compared with visible textual manipulations~\cite{Cui2024RobustnessCVPR,Xie2025ChainAttackCVPR, Liu2024IJCAISafety, Li2024AchillesECCV}.

Recently, many approaches are proposed to make VLMs robust against these adversarial image perturbations, including adversarial training~\cite{Schlarmann2024RobustCLIPICML,Zhang2024AdvPTECCV,Zhou2024FewShotAPTNeurIPS,Mao2023TeCoAICLR}, adversarial prompt tuning~\cite{Zhang2024AdvPTECCV,Zhou2024FewShotAPTNeurIPS} and test-time tuning~\cite{Wang2025TAPTCVPR,Sheng2025RTPTCVPR}.
Despite steady progress, these defense mechanisms have some practical limitations. 
(1) \textbf{Gradient access}: first-order methods~\cite{Mao2023TeCoAICLR,Schlarmann2024RobustCLIPICML,Zhang2024AdvPTECCV,Zhou2024FewShotAPTNeurIPS} rely on access to input gradients, which is infeasible for many closed-source VLMs where only output tokens can be obtained via APIs.
(2) \textbf{Generalization to unseen attacks}: training-time approaches~\cite{Mao2023TeCoAICLR,Schlarmann2024RobustCLIPICML,Zhang2024AdvPTECCV,Zhou2024FewShotAPTNeurIPS} often use adversarial perturbations of specific types, such as $\ell_\infty$-bounded ones. This limits their generalizability, leaving models vulnerable to unseen perturbations or tasks~\cite{Schlarmann2024RobustCLIPICML,Zhang2024AdvPTECCV,Zhou2024FewShotAPTNeurIPS}.
(3) \textbf{Restrictive scope}: some methods~\cite{Zhang2024AdvPTECCV,Zhou2024FewShotAPTNeurIPS,Wang2025TAPTCVPR,Sheng2025RTPTCVPR} rely on specific training objectives or architectural designs of the backbone model, making them unsuitable for general-purpose applications.
(4) \textbf{Trade-offs with clean inputs}: several methods~\cite{Schlarmann2024RobustCLIPICML,Mao2023TeCoAICLR} improve robustness at the expense of degrading performance on clean inputs~\cite{Tsipras2019TradeoffICLR}, reducing the overall utility of robust models.


To address the concerns pointed above, we propose \textbf{Paraphrase-Decomposition-Aggregation (PDA)}, a generic, training-free framework that only needs black-box access to defend the VLMs against adversarial image perturbations.
Our design is inspired by randomized smoothing~\cite{cohen2019certified} and its adaptation to large language models (LLMs)~\cite{Robey2023SmoothLLM,Ji2024NAACL_SmoothingSurvey}, where several inputs are sampled in the neighborhood of each input data and the corresponding outputs are aggregated to ensure stabilized decisions under adversarial attacks.
In the context of VLMs, many adversarial attacks~\cite{Yin2023VLAttackNeurIPS,AttackVLM} exploit the similarity between the image input and the text input to craft coordinated adversarial perturbations, suggesting that searching a local neighborhood in the high-dimensional text space can help recover semantics consistent with the image. 
Building on this intuition, PDA freezes model parameters and operates purely at test time via three steps: (i) \textbf{Paraphrase} the user query into multiple semantically equivalent views to exploit linguistic redundancy; (ii) \textbf{Decompose} the task into verifiable atomic questions that expose stable, image-aligned evidence; and (iii) \textbf{Aggregate} the answers with agreement-/confidence-aware voting to suppress adversarially biased views.

\begin{wrapfigure}{r}{0.53\linewidth} 
  \centering
  \vspace{-10pt}
  \includegraphics[width=\linewidth]{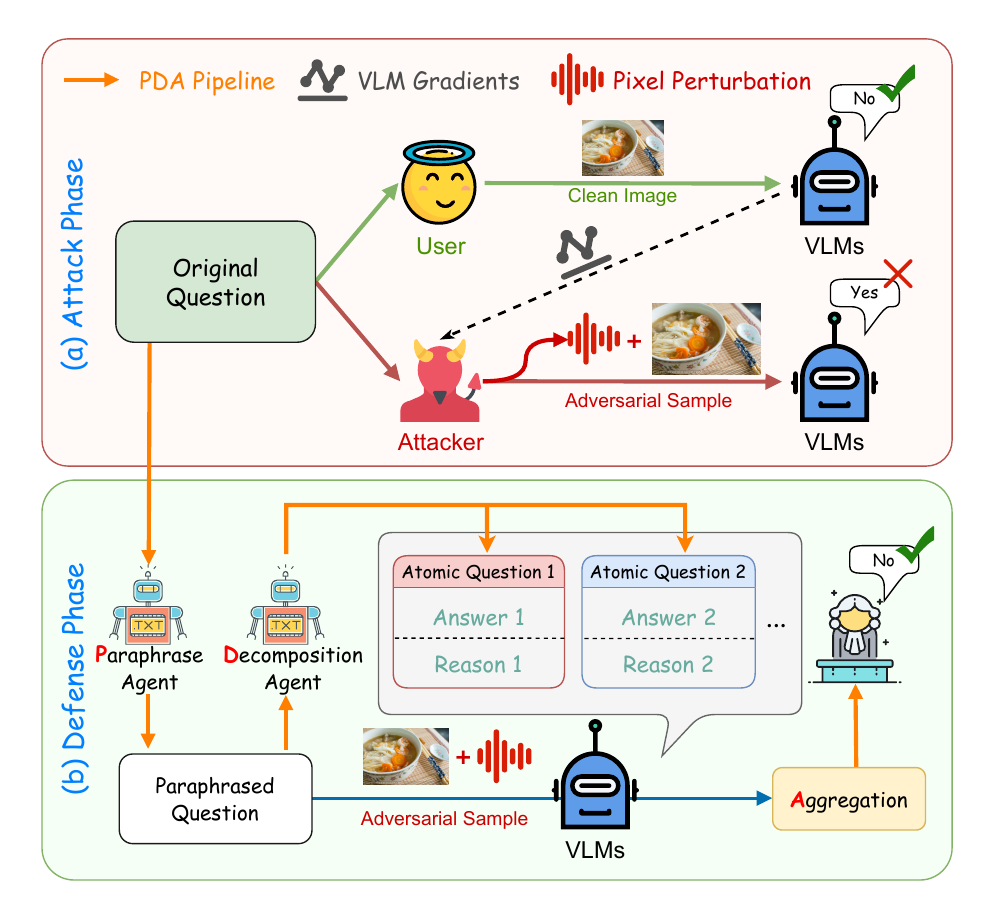}
  \vspace{-16pt}
  \caption{\textbf{Overview of the threat model and PDA.} 
  \emph{(a) Attack phase:} an adversary adds pixel-level perturbations guided by VLM gradients, causing the model to flip its answer under the original question. 
  \emph{(b) Defense phase:} our training-free, text-side Paraphrase–Decomposition–Aggregation pipeline, requires only black-box access, and suppresses adversarially biased views by cross-checking stable visual evidence.}
  \label{fig:intro_overview}
  \vspace{-20pt}
\end{wrapfigure}

Our proposed framework PDA assumes black-box access to VLMs, it is agnostic to model architectures, tasks and perturbations types.
Therefore, it can be applied to various applications in a plug-and-play manner.
To assess generality and practicality, we evaluate the framework across common VLM use cases, including visual question answering, zero-shot classification, and image captioning, covering multiple model families and parameter scales. The extensive results demonstrate effectiveness of PDA: it yields consistent gains under adversarial image attack while largely preserving the performance with clean inputs. Our contributions are summarized as follows:
\begin{itemize}[leftmargin=*]
  \item We propose \textbf{Paraphrase-Decomposition-Aggregation (PDA)} framework, which supports black-box access and is training-free. It establishes a generic paradigm to defend VLMs against adversarial image perturbations.
  \item We validate the effectiveness of PDA across a broad spectrum of applications, including different tasks and different model families with varying number of parameters and access controls. The extensive results show consistent robustness gains from PDA against various adversarial image perturbations with small performance degradations in clean inputs, demonstrating the broad applicability and effectiveness of our proposed framework.
  \item We further introduce several PDA variants that target different efficiency requirements and retain most of the robustness improvements. They substantially reduce the inference cost, yielding favorable accuracy–latency trade-offs across tasks and backbones.
\end{itemize}

%% file: sec/2_related_work.tex
\section{Related Work}
\label{sec:rw}

\noindent\textbf{Attacks against Vision-Language Models.}
Similar to the uni-modal models~\cite{szegedy2013intriguing, goodfellow2015explaining, madry2018towards, athalye2018obfuscated, croce2020reliable}, a growing body of evidence~\cite{Cui2024RobustnessCVPR, AttackVLM, Yin2023VLAttackNeurIPS, Zhang2025AnyAttackCVPR, Xie2025ChainAttackCVPR}  shows that Vision-Language Models (VLMs) are vulnerable to many kinds of adversarial attacks. 
Small and human-imperceptible adversarial image perturbations can cause significant performance degradation for VLMs across multiple tasks. 
Adversarial attacks on pre-trained VLMs can be broadly categorized by the attacker’s knowledge and the way multimodal inputs are perturbed. 
In the white-box setting, adaptations of classical image attacks such as Fast Gradient Sign Method (FGSM)~\cite{goodfellow2015explaining}, Projected Gradient Descent (PGD)~\cite{madry2018towards} and AutoAttack~\cite{croce2020reliable} can be applied directly to the visual encoders. 
Under black-box constraints, AttackVLM~\cite{AttackVLM} systematically evaluates transfer- and query-based targeted attacks on a wide range of instruction-tuned VLMs. Follow-up black-box methods further strengthen this threat surface by using pre-trained VL encoders to craft image–text pair perturbations and improve transferability and scalability, including VLATTACK~\cite{Yin2023VLAttackNeurIPS}, Chain of Attack~\cite{Xie2025ChainAttackCVPR}, and AnyAttack~\cite{Zhang2025AnyAttackCVPR}.
All these observations underscore the increasing security concerns of adversarial visual inputs.
\noindent\textbf{Defenses for Vision-Language Models.}
A large body of work aims to improve the adversarial robustness of CLIP-based VLMs, and existing defenses can be broadly divided into \emph{training-time adversarial tuning} and \emph{test-time adaptation}. At training time, adversarial training methods such as TeCoA~\cite{Mao2023TeCoAICLR} and FARE~\cite{Schlarmann2024RobustCLIPICML} adversarially fine-tune the CLIP vision encoder with text-guided contrastive or robust embedding objectives so that zero-shot robustness transfers to downstream VLMs. 
Beyond modifying the backbone, adversarial prompt tuning methods~\cite{Zhang2024AdvPTECCV, Zhou2024FewShotAPTNeurIPS, Yu2024TGAZSRNeurIPS, wang2024one} keep CLIP encoders frozen and instead learn robust text prompts or attention modules.
Overall, these training-time approaches defensively reshape the visual or language branch but still rely on generating adversarial examples and substantial offline optimization, which limits their applicability when data are restricted.

To avoid the need for additional data and retraining, recent work has begun to explore \emph{test-time defenses} that adapt prompts at inference. Test-Time Adversarial Prompt Tuning (TAPT)~\cite{Wang2025TAPTCVPR} learns bi-modal defensive prompts per input by minimizing multi-view entropy and aligning clean–adversarial feature distributions. R-TPT~\cite{Sheng2025RTPTCVPR} refines this paradigm with a reformulated entropy objective and reliability-weighted aggregation over augmented views, and C-TPT~\cite{Yoon2024CTPTICLR} further calibrates prompt updates using text-feature dispersion. Beyond prompt tokens, efficient test-time adaptation such as TDA~\cite{Karmanov2024TDACVPR} uses training-free dynamic adapters and key–value caches to reduce overhead under distribution shift. Beyond tuning-based defenses, several studies instead analyze how architectural choices and simple prompt patterns themselves affect robustness~\cite{Bhagwatkar2024ArchPromptRobustness,Nirala2023OVC}. Taken together, these test-time methods rely on adapter optimization over augmented views, are mostly evaluated on CLIP-style zero-shot classification, and still assume white-box gradients or internal representations, which limits their applicability in realistic black-box VLM deployments.

\noindent\textbf{Randomized Smoothing.}
Randomized smoothing provides instance-wise robustness certificates by predicting with a noise-perturbed ensemble and assigning the class by majority vote under the noise distribution. Classical work establishes tight $\ell_2$ guarantees for Gaussian smoothing and ImageNet-scale certification~\cite{Cohen2019RS}, and is further refined by adversarially trained smoothed classifiers, radius-maximizing objectives, and generalized smoothing measures~\cite{Salman2019SmoothAdv,Zhai2020MACER,Yang2020RSShapes}.
Originally developed for image classification, randomized smoothing has since been extended to large language models. For example, SmoothLLM~\cite{Robey2023SmoothLLM} perturbs prompts at the character level to stabilize large language models under jailbreak attempts, and SemanticSmooth~\cite{Ji2024NAACL_SmoothingSurvey} aggregates predictions over semantically transformed prompts to defend against stronger prompt-level attacks. More recently, smoothing ideas have been explored for vision--language models. PromptSmooth learns zero- or few-shot textual prompts so a fixed Med-VLM remains accurate under Gaussian image noise~\cite{Hussein2024PromptSmooth}, and Open-Vocabulary Certification accelerates CLIP-style certification via incremental smoothing and caching for novel prompts~\cite{Nirala2023OVC}. However, these methods mostly smooth the visual or embedding space for specific architectures and classification tasks, without reshaping the full multimodal decision process or generated outputs, motivating randomized-smoothing-style defenses that act directly at the VLM decision level in more general black-box settings.

%% file: sec/3_methodology.tex
\section{Methodology}
\label{sec:method}
\subsection{Preliminary}

\noindent\textbf{Formulation.} We use function $f_\theta(x; t) \rightarrow y$ to represent a Vision-Language Model (VLM) which is parameterized by $\theta$ and maps an image \( x \) and a text input \( t \) to an output \( y \).
Here, the text input \( t \) is typically a question or directive related to the image $x$, and the output \( y \) is the corresponding answer generated by the VLM. 
A correct and coherent response should capture the information from both the visual input and the textual input.

\noindent\textbf{Adversarial Perturbation.}
Adversarial attacks on the visual input aim to perturb the image \( x \) to make the VLM generate incorrect outputs. It is formulated as an optimization problem to minimize a loss objective function \( \mathcal{L} \) on the perturbation \( \delta \) with a constraint on its size, usually based on its $l_p$ norm: 
\begin{equation} \label{eq:problem}
    \delta^* = \arg \max_{\delta} \mathcal{L}(f_\theta(x + \delta; t), y) \quad \text{subject to} \quad \| \delta \|_p \leq \epsilon
\end{equation}

In this study, we assume a white-box threat model, meaning that the attacker has full access to the model, including its architecture, and parameters. 
This assumption allows us to evaluate our defense strategy under the most challenging conditions. 
The white-box attackers have access to the gradient of the loss with respect to the input perturbation, i.e. $\nabla_\delta \gL$, which enables the use of gradient-based adversarial attack methods such as FGSM and PGD to efficiently approximate the optimal perturbation $\delta^*$ in Problem~(\ref{eq:problem}). 

\noindent\textbf{Defense Based on Text Perturbations.}
As multi-modal models, the output of VLMs depends on both visual and textual inputs.
This dependency makes it possible to manipulate textual inputs to counteract the effects of adversarial image perturbations.
Inspired by randomized smoothing~\cite{Cohen2019RS}, we expect that adversarial images often lie within the spike regions close to the decision boundary.
By introducing small perturbations to the textual inputs without significantly altering their semantic meanings, we can aggregate the corresponding outputs to obtain smoother and thus more robust results.


To formally define these text perturbations, we introduce a perturbation function \( \mathcal{T}(\cdot) \), which maps the original text input \( t \) to the perturbed text \( t' = \mathcal{T}(t) \).
In this context, $\gT$ paraphrases the original text $t$, aiming to alter the literal expression of the text $t$ while maintaining its underlying semantic meanings.
Compared with existing works~\cite{Robey2023SmoothLLM, Ji2024SemanticSmooth} which apply randomized smoothing in uni-modal large language models, we consider a richer family of text perturbations.
In addition to character-level edits and synonym replacements, we include sentence-level semantic transformations in $\gT$, including sentence rephrasing by grammatical restructuring and decomposition of a single complex query into several simpler atomic questions, as shown in Figure~\ref{fig:example}.
These transformations allows us to exploit more high-level, complex and interpretable features extracted by VLMs, which is robust against small and imperceptible image perturbations, since extensive literature~\cite{tsipras2018robustness, ilyas2019adversarial, salman2020adversarially, wang2020high} demonstrate that adversarial perturbation usually alters high-frequency, low-level and uninterpretable features.


\begin{figure*}[t]
  \centering
  \includegraphics[width=\linewidth]{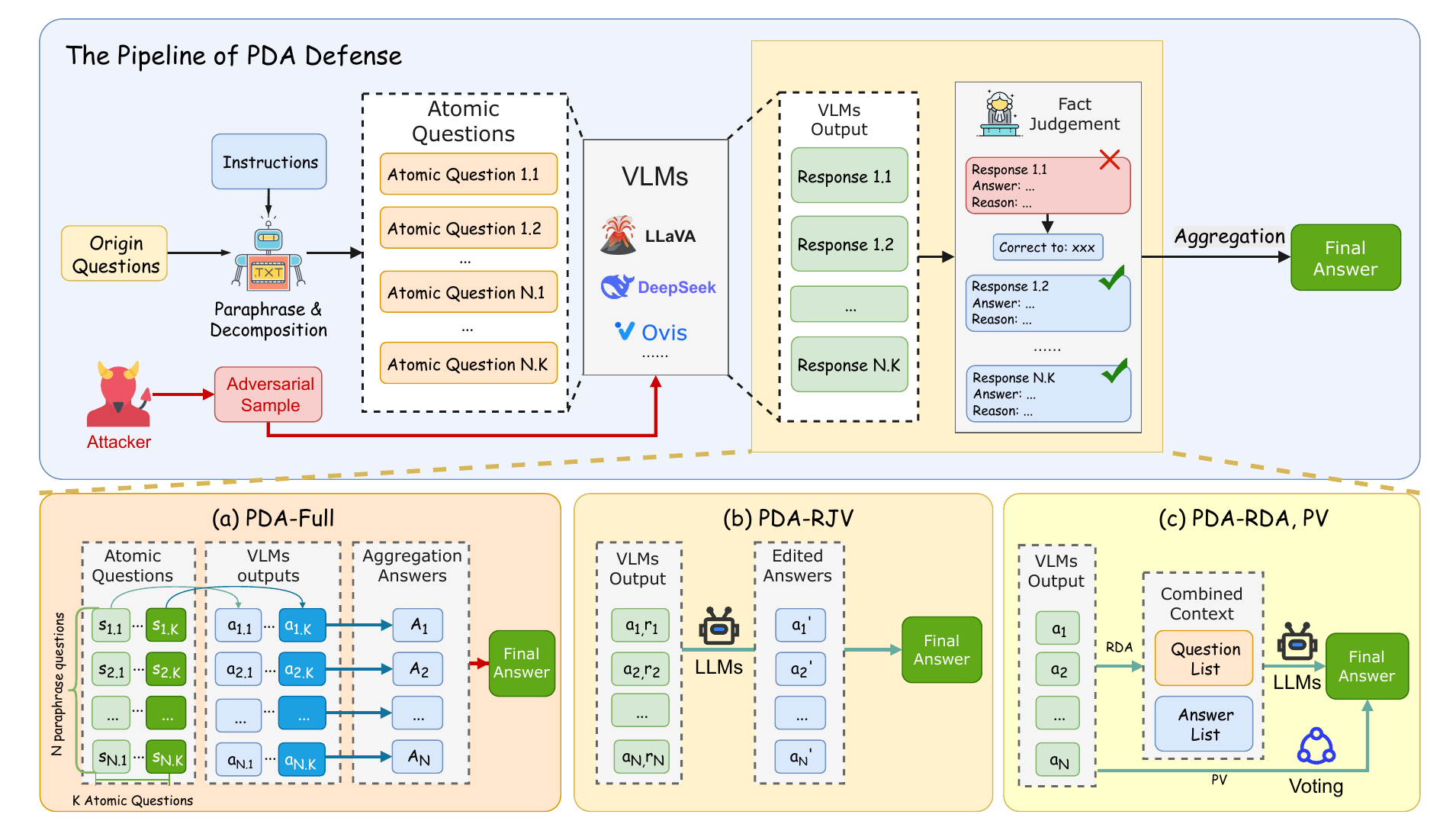}
  \caption{\textbf{Pipeline of the proposed PDA defense and its variants.}
  Given an image $x$ and a query $t$, the \emph{Paraphrase Agent} produces semantically equivalent prompts.
  Each prompt is factorized by the \emph{Question Decomposer} into atomic questions that are individually answered by the base VLM.
  The \emph{Answer Aggregation Agent} performs confidence- and consistency-aware fusion to yield the final prediction. The bottom panels illustrate four PDA variants: (a) \textbf{PDA-Full}, (b) \textbf{PDA-RJV}, (c) \textbf{PDA-RDA} and \textbf{PDA-PV}.}
  \label{fig:pipeline}
\end{figure*}

\subsection{PDA Defense Framework}
\label{subsec:pda}

Based on the motivations above, we introduce \textit{Paraphrase-Decomposition-Aggregation (PDA)} as a generic, training-free defense framework that only needs black-box access to improve the robustness of VLMs against adversarial image perturbations.
As the name indicates, PDA consists of three stages: (1) paraphrase the text inputs to obtain several semantically similar and literally distinct prompts; (2) decompose each prompt into a sequence of several atomic questions; (3) feed atomic questions to VLMs and aggregate the corresponding output to obtain reliable final answers.
The overall pipeline is demonstrated in Figure~\ref{fig:pipeline} and we discuss the details of each stage below.



\noindent {\textbf{Step I: Paraphrase.}}
Paraphrase aims to explore the ``neighborhood'' in the semantic space of the original textual input $t$.
In this stage, we employ a paraphrase agent $\gT_1$, usually a large language model (LLM), to generate a finite set of candidate paraphrases:
\begin{equation}
\gS \;=\; \{t_1,\dots,t_n\}, \qquad t_i \sim \gT_1(t;\phi) \label{eq:phase1}
\end{equation} 
where $\gT_1(t; \phi)$ denotes the paraphrase generation distribution and $\phi$ denotes the prompt.
The prompt contains the detailed information for paraphrase, which controls the diversity of the output texts.
A higher degree of diversity encourages more aggressive rephrasings and structural variants, while a more deterministic outputs keeps the paraphrases closer to the original wording.

The set $\gS$ we obtain in (\ref{eq:phase1}) is the collection of admissible text transformations applied to the original textual input. 
Therefore, the responses from VLMs for the paraphrased texts in $\gS$ will establish an \emph{Expectation over Text Transformations} (EoTT) estimator for robust inference, which is analogous to the Expectation over Transformations (EOT) principle~\cite{athalye2018obfuscated,Horvath2022DRS,Schuchardt2023LocalizedRS,Ugare2024IRS} commonly used in adversarial vision, but instantiated in the textual domain.

\noindent {\textbf{Step II: Decomposition.}}
The success of chain-of-thought prompting~\cite{wei2022chain} indicates that it is more reliable to solve a problem as a series of intermediate steps than directly giving a final answer.
Similarly, we decompose the paraphrased question obtained in Step I into a sequence of simpler and atomic questions for more reliable inference.
By exploring the answer at the level of finer-grained evidence instead of a single holistic answer, we explicitly make the reasoning process more interpretable, which is shown beneficial to robustness~\cite{Tsipras2019TradeoffICLR, ilyas2019adversarial}.
In the context of VLMs, we reduce a potentially complex, multi-hop query into a sequence of small atomic questions, each targeting a single visual fact like the presence of an object, its attributes, spatial relations, counts, or simple comparisons.

Formally, for each paraphrase $t_i$ obtained in Step I, we employ a decomposition agent $\gT_2$, usually a large language model (LLM), to explicitly construct a finite sequence of atomic questions:
\begin{equation}
\mathcal{S}_i \;=\; \{s_{i,1},\dots,s_{i,k_i}\} \;\sim\; \gT_2(t_i, \varphi),
\end{equation}
where $s_{i, j}$ is an atomic question and $\varphi$ is the prompt containing the decomposition instructions for generating the sequence. 
The prompt $\varphi$ should ensure that the decomposition is \textit{sufficient} and \textit{faithful}.
Specifically, when querying the VLMs with the sequence of atomic questions, the responses $\{a_{i,j} = f_\theta(x; s_{i, j})\}_{j = 1}^{k_i}$ should be sufficient to construct a meaningful answer to question $t_i$.
In addition, each atomic question should refer only to evidence that is in principle available in the image $x$, avoiding the introduction of extraneous assumptions or off-image knowledge.
We provide the prompt example to fulfill these two requirements in the supplemental materials.



\noindent {\textbf{Step III: Aggregation.}}
In Step I and Step II, we do not use the image for paraphrase and decomposition, since the potential adversarial image perturbation may affect the construction of atomic questions.
In Step III, we feed these sequences of atomic questions into the VLM and obtain the corresponding responses.
We use the evidence set 
\(
\mathcal{B}_i \;=\; \{(s_{i,j}, a_{i,j})\}_{j=1}^{k_i},
\)
to represent the question-answer pairs for paraphrase $t_i$, and then entire collection for different paraphrases is $\gB = \cup_{i = 1}^n \gB_i$. 
We use the aggregate agent in Step III to obtain the final results through two-level aggregation: it first summarizes $\mathcal{B}_i$ into a single candidate answer for $t_i$, then summarizes the answers for different paraphrases to generate the final results.

For structured tasks with a closed candidate set $\mathcal{Y}$ (e.g., multiple choices in VQA or image classification), we instantiate this agent as a lightweight decision head that maps each evidence set to a label
\(
g(\mathcal{B}_i) \;=\; \widehat{y}_i   \in \mathcal{Y},
\)
where $g$ can be implemented as a simple voting rule, such as simple majority, over the sub-answers in $\mathcal{B}_i$. 
The same rule can be applied to aggregate results from $\gB_i$ to obtain the final results for $\gB$.
For open-form tasks such as image captioning, we use an LLM $h$ as the aggregation agent to directly generate the summary $\widehat{a} = h(x, t, \mathcal{B}, \Phi)$ as the final results.
Here, $\Phi$ is the prompt describing how the responses in $\gB$ are organized and potential task-specific information. In this way, Step III concludes the pipeline of PDA, establishing a generic, training-free and black-box defense framework. The pseudo-code is demonstrated in Algorithm~\ref{alg:pda}.

\begin{algorithm}[t]
\caption{PDA-VLM: training-free PDA wrapper for robust inference with a frozen VLM $f_\theta$}
\label{alg:pda}
\begin{algorithmic}[1]
\REQUIRE Image $x$, query $t$; paraphrase budget $n$;  
         task type (structured with candidate set $\mathcal{Y}$, or open-form).
\STATE \textbf{// Step I: Paraphrase}
\STATE Sample candidate paraphrases $\{t_i\}_{i=1}^{n} \leftarrow \gT_1(t;\phi)$.
\FOR{each retained paraphrase $t_i$}
  \STATE \textbf{// Step II: Decomposition}
  \STATE $\mathcal{S}_i \leftarrow \gT_2(t_i, \varphi)$ \COMMENT{obtain atomic questions}
  \FOR{each $s_{i,j} \in \mathcal{S}_i$}
    \STATE $a_{i,j} \leftarrow f_\theta(x; s_{i,j})$ \COMMENT{answer atomic question}
  \ENDFOR
  \STATE $\mathcal{B}_i \leftarrow \{(s_{i,j}, a_{i,j})\}_{j=1}^{k_i}$ \COMMENT{evidence for paraphrase $t_i$}
\ENDFOR
\STATE \textbf{// Step III: Aggregation}
\IF{task is structured with closed candidate set $\mathcal{Y}$}
  \FOR{each paraphrase $t_i$}
    \STATE $\widehat{y}_i \leftarrow g(\mathcal{B}_i) \in \mathcal{Y}$ \COMMENT{e.g., voting over mapped sub-answers}
  \ENDFOR
  \STATE Compute evidence scores $E(y) \leftarrow \sum_{i} \mathbb{I}[\widehat{y}_i = y]$ for all $y \in \mathcal{Y}$.
  \STATE \textbf{return} $\widehat{y}^\star \leftarrow \arg\max_{y \in \mathcal{Y}} E(y)$
\ELSE
  \STATE $\mathcal{B} \leftarrow \bigcup_{i} \mathcal{B}_i$ \COMMENT{collect all atomic question--answer pairs}
  \STATE $\widehat{a} \leftarrow h(x, t, \mathcal{B})$ \COMMENT{text-only LLM aggregates into a free-form answer}
  \STATE \textbf{return} $\widehat{a}$
\ENDIF
\end{algorithmic}
\end{algorithm}

\subsection{Budgeted PDA Variants.}

PDA establishes a generic three-step framework for robust inference against adversarial image perturbations.
Similar to randomized smoothing, there is an effectiveness-complexity trade-off under PDA framework.
In this context, we instantiate four pipelines that differ in where aggregation occurs and how paraphrasing and decomposition are coupled.
We summarize the complexity of these variants in Table~\ref{tab:complexity}.
In the supplemental materials, we provide some concrete examples to demonstrate their workflows.

\begin{wraptable}{r}{0.48\linewidth}
\centering
\vspace{-16pt}
\footnotesize
\setlength{\tabcolsep}{4pt}
\renewcommand{\arraystretch}{1.08}
\caption{Computation budget by stage (LLM: text-only; VLM: vision--language). Here $M$ is the number of retained paraphrases and $K$ is the number of atomic questions per paraphrase (generated within a single VLM call per paraphrase).} 
\label{tab:complexity}
\resizebox{\linewidth}{!}{%
\begin{tabular}{@{}l c c cc@{}}
\toprule
\textbf{Variant} &
\textbf{Paraphrasing} & \multicolumn{2}{c}{\textbf{Decomposition}} &
\textbf{Aggregation} \\
 & \textbf{LLM} & \textbf{VLM} & \textbf{LLM} & \textbf{LLM} \\
\midrule
\textbf{PDA-Full} & 1 & $N\!\times\!K$ & $N\!\times\!K + N$ & 1 \\
\textbf{PDA-RJV}  & 1 & $N$ & $N$                 & $0$ \\
\textbf{PDA-RDA} & 1 & $N$ & 0             & 1 \\
\textbf{PDA-PV} & 1 & $N$ & 0             & 0 \\
\bottomrule
\end{tabular}
}
\vspace{-12pt}
\end{wraptable}
\noindent\textbf{(1) PDA-Full.}
This variant executes the full chain. A single LLM is employed to paraphrase the original question $t$ into $N$ paraphrases, and each paraphrase into up to $K$ atomic questions. The atomic questions are fed to VLMs together with the potentially adversarial image input. The VLM responses are then filtered at the probe level based on their consistency. The LLM is employed again in the final aggregation step to produce the decision.


\noindent\textbf{(2) PDA-RJV (Reason--Judge--Vote).} This variant skips decomposition step. The VLM directly answers each of the $N$ paraphrases (but with a brief rationale). A small LLM is then employed to \emph{judge} the $N$ paraphrase-level answers one by one and summarize the final decision based on the weighted vote across these paraphrase-level answers.
Compared with PDA-Full, PDA-RJV employs LLM as judges and apply voting aggregation on the paraphrase level, cutting a significant number of LLM calls while letting the judge correct possible incorrect answers using cross-paraphrase evidence.


\noindent\textbf{(3) PDA-RDA (Reason--Direct--Aggregate).} This variant skips the decomposition step and VLM answers each of the $N$ paraphrases with a brief rationale. Compared with PDA-RJV, PDA-RDA directly employs an LLM to jointly reviews all $(\text{paraphrase}, \text{answer}, \text{rationale})$ tuples and outputs the final decision \emph{in one shot} by a single call in aggregation  step. It replaces $N$ paraphrase-level judgements with one global pass, reducing the number of LLM calls while assessing the VLM responses from a global perspective.


\noindent\textbf{(4) PDA-PV (Paraphrase--Vote).} This variant targets the structured tasks with a closed candidate set. It skips the decomposition step and further integrate the paraphrase and aggregation steps. It generate paraphrases that are not necessary semantically equivalent but logically equivalent to pick the correct answer. For example, paraphrasing ``t-shirt / jeans'' to ``upper wear / lower wear'' is not semantically equivalent, but such paraphrase preserves the correct choice. Because of larger semantic perturbations, PDA-PV increases the diversity of paraphrases. It reduces the number of LLM calls to $N$ paraphrases and one voting aggregation.


%% file: sec/4_exp.tex
\section{Experiments}
\label{sec:exp}

\newcommand{\cmark}{\checkmark}
\newcommand{\xmark}{\phantom{\checkmark}}
\newcommand{\eps}{\varepsilon}
\newcommand{\pgd}{\textsc{PGD}\xspace}
\newcommand{\adv}{\textsc{Adv}\xspace}
\newcommand{\clean}{\textsc{Clean}\xspace}
\newcommand{\eot}{\textsc{EOT}\xspace}
\newcommand{\ran}{\textsc{RA}\xspace}

\newif\ifshowrevision
\showrevisionfalse

\ifshowrevision
    \newcommand{\rev}[1]{\textcolor{blue}{#1}}
\else
    \newcommand{\rev}[1]{#1}
\fi


\subsection{Experimental Setup Overview}

\noindent \textbf{Datasets and Metrics.}
We adopt VQA-v2~\cite{Goyal2017VQAv2}, ImageNet\textnormal{-}D~\cite{Zhang2024ImageNetD}, and MS~COCO~\cite{Chen2015COCOCaptions} with their community-standard metrics to target three complementary facets of robustness.
VQA-v2 is used to probe open-ended multimodal reasoning under reduced language priors, so improvements under attack are less attributable to textual shortcuts.
ImageNet\textnormal{-}D evaluates recognition under controlled generative shifts that preserve category semantics while perturbing nuisance factors, offering a focused stress test of visual backbones and VL pipelines.
For COCO captioning, following recent robustness evaluations~\cite{AttackVLM}, we report \emph{CLIPScore} to quantify the similarity between generated captions and ground-truth descriptions.
Accordingly, we use \emph{VQA Accuracy} on VQA-v2, \emph{Top-1 Accuracy} on ImageNet\textnormal{-}D, and \emph{CLIPScore} on COCO, aligning our protocol with prevailing practice and enabling direct comparability with prior robustness studies.

\noindent\textbf{Models.}
Primary evaluations use \emph{LLaVA-1.5-7B} and \emph{LLaVA-1.5-13B}~\cite{Liu2024LLaVA15}, widely adopted LVLMs with strong instruction-following capability.
To test plug-and-play generality across architectures and parameter scales, we additionally evaluate \emph{DeepSeek-VL-1.3B}~\cite{Lu2024DeepSeekVL}, \emph{InternVL3-2B}~\cite{InternVL3_2025}, and \emph{Ovis2-4B}~\cite{Lu2024Ovis}.
These families differ in visual encoders, connector designs, and training recipes, providing a diverse set of backbones to examine whether a training-free text-side wrapper transfers without re-training.

\noindent\textbf{Threat model and Baselines.}
We evaluate robustness under both white-box and black-box adversarial settings.
For white-box evaluation, we adopt \pgd as the standard attack~\cite{madry2018towards}, and further consider adaptive attacks based on expectation over transformations (\eot), which explicitly optimize over PDA's stochastic paraphrase and decomposition pipeline.
For black-box evaluation, we include representative transfer- and query-based attack settings, including the transfer-based protocol in~\cite{Cui2024RobustnessCVPR}, \emph{AttackVLM}, and \emph{AnyAttack}~\cite{AttackVLM,Zhang2025AnyAttackCVPR}.
As text-side randomized-smoothing baselines, we include \emph{SmoothLLM} with Swap/Insert/Patch perturbations~\cite{Robey2023SmoothLLM} and \emph{SemanticSmooth} with paraphrase/summarise transformations~\cite{Ji2024SemanticSmooth}.
As training-time robustification references, we evaluate released checkpoints from \emph{TeCoA} and \emph{FARE}~\cite{Mao2023TeCoAICLR,Schlarmann2024RobustCLIPICML}.
Our method (\emph{PDA}) is training-free and operates solely on the textual interface; all methods are compared under the same threat model and data splits for fairness.

\noindent\textbf{PDA Configurations.}
Unless otherwise specified in the corresponding subsection, the primary comparison results use the \textbf{PDA-Full} pipeline with \texttt{GPT-4.1-mini} as the default LLM agent.
For VQA-v2 and ImageNet-D, PDA-Full uses $5$ paraphrases per query, and each paraphrase is further decomposed into $3$ atomic sub-questions, yielding $15$ sub-questions in total.
For each paraphrase, the answers to its three sub-questions are aggregated by the LLM into one paraphrase-level answer, and the final prediction is obtained by majority voting over the five paraphrase-level answers.
For COCO captioning, we use $2$ paraphrases and $5$ atomic questions to better probe object/attribute grounding while keeping decoding cost moderate.
SmoothLLM and SemanticSmooth both aggregate $10$ randomized views by majority/score voting.
For evaluation of TeCoA and FARE, the evaluation is run by direct inference under \pgd attack with the public weights provided by~\cite{Schlarmann2024RobustCLIPICML}.

\rev{
we also study several lighter-weight variants in the ablation studies section.
\emph{PDA-RJV} keeps the same paraphrase width ($5$) but removes decomposition: for each paraphrase, the victim VLM outputs an answer and a rationale, and the LLM judges each paraphrase independently; the final answer is determined by majority voting over the five judgments.
\emph{PDA-RDA} also removes decomposition while keeping five paraphrases, but instead of judging each paraphrase separately, it feeds all five paraphrase--answer--rationale tuples to the LLM in a single pass to directly produce the final prediction.
\emph{PDA-PV} is the lightest configuration: it uses five logically equivalent paraphrases, removes both decomposition and LLM-based aggregation, queries the VLM once per paraphrase without requiring rationales, and obtains the final prediction by majority voting over the five VLM answers.
Unless explicitly specified in the corresponding subsection, the main comparison tables use PDA-Full, while variant-specific results are reported separately in the ablation studies.
}

\subsection{Robustness against White-box Attacks}
\label{subsec:overall_eval}

\begin{figure*}[t]
  \centering
  \includegraphics[width=0.98\linewidth]{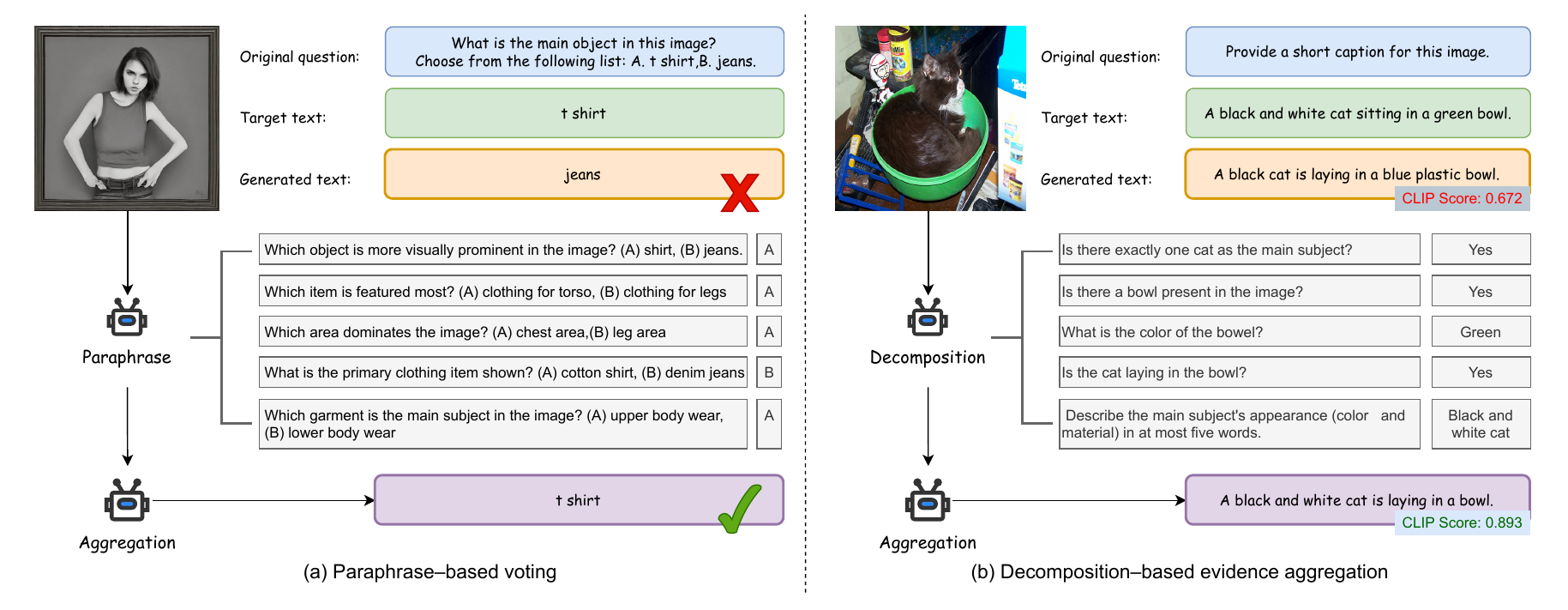}
  \caption{\textbf{Qualitative effect of PDA.}
(a) The base VLM is attacked into choosing ``jeans'' in a two-way ``t shirt vs.\ jeans'' question, whereas PDA paraphrases the query into targeted checks and recovers the correct ``t shirt'' label.
(b) For a cluttered scene with a black-and-white cat in a green bowl, PDA decomposes the caption prompt into factual sub-questions and aggregates the answers into a faithful caption.}
  \label{fig:example}
\end{figure*}

\noindent \textbf{Standard PGD.} We start with PGD as the white-box attack, using a perturbation budget $\epsilon = 2 / 255$. Across the three tasks and both LLaVA scales, \emph{PDA} markedly increases adversarial accuracy while keeping clean accuracy essentially unchanged (Table~\ref{tab:main}).
On VQA-v2, ADV rises to 0.754 (7B) and 0.787 (13B); on ImageNet\textnormal{-}D we observe higher clean accuracy alongside large adversarial gains (e.g., 7B clean $0.758$ vs.\ $0.625$).
This behavior matches the mechanism of PDA’s \emph{paraphrase--/decompose--aggregate} pipeline: atomic questions cross-check stable visual evidence and aggregation self-corrects spurious predictions that would otherwise contaminate both robust and nominal outputs.
The qualitative examples in Fig.~\ref{fig:example} further illustrate this effect: for classification, PDA overturns an incorrect ``jeans'' decision by paraphrasing the main-object query into checks on prominence, region, and garment type and then voting for ``t shirt''; for captioning, PDA decomposes the prompt into factual questions (object count, color, pose) and aggregates the answers into a more faithful caption with a higher CLIP score.
Additional qualitative cases are provided in the supplemental materials.

\input{sec/tables/overall}

Compared with text-side randomized-smoothing baselines, SmoothLLM and SemanticSmooth yield only modest robustness changes and sometimes degrade clean performance; this is consistent with recent observations that character-level noise or shallow synonym substitutions often induce small movements in the model’s semantic space for modern instruction-tuned LLMs, limiting downstream effect when applied naively to VLM prompts.
Finally, training-time defenses typically report stronger robustness but can exhibit a clear robustness--utility trade-off on clean inputs.
In contrast, PDA achieves large adversarial accuracy improvements without any additional training or parameter updates and requires only black-box access at inference, providing a low-cost and architecture-agnostic alternative to fine-tuning--based defenses.

\noindent \textbf{Larger Perturbation Budgets.}
We further stress-test PDA under more challenging white-box settings that go beyond the default \pgd attack, focusing on both larger perturbation budgets.
\rev{We evaluate robustness by sweeping the \pgd perturbation budget $\epsilon\in\{2,4,6,8\}/255$.
This experiment tests whether the robustness gains of PDA are confined to a narrow perturbation regime or persist as the perturbation magnitude increases.
As shown in Table~\ref{tab:stronger_whitebox}(a), the no-defense baseline degrades steadily as $\epsilon$ grows on both ImageNet\textnormal{-}D and VQA-v2, whereas PDA remains comparatively stable across budgets.
These trends indicate that PDA is not tuned to a single attack budget and continues to provide substantial robustness gains even under stronger white-box perturbations.}

\noindent \textbf{Adaptive Attacks.}
In addition to larger adversarial perturbations, we further evaluate robustness against adaptive attacks that explicitly target the stochastic components of PDA, \rev{since PDA relies on stochastic paraphrase and decomposition, a vanilla white-box \pgd attacker that differentiates through only a single sampled view may be mismatched to the defense.
Therefore, we adopt expectation over transformations (EoT)~\cite{athalye2018obfuscated} and optimize the expected loss over PDA's stochastic components:
\begin{equation}
\max_{\delta}\ \mathbb{E}_{p,d}\!\left[\mathcal{L}\!\left(f_{\text{PDA}}(x+\delta; p,d), y\right)\right],
\end{equation}
where the expectation over sampled paraphrases $p$ and decompositions $d$ is approximated by Monte Carlo samples during optimization.
We evaluate three variants that isolate different sources of randomness: \emph{EOT-P} (expectation over paraphrases), \emph{EOT-D} (expectation over decompositions), and \emph{EOT-PD} (joint expectation over both).}

\rev{As shown in Table~\ref{tab:stronger_whitebox}(b), these adaptive attacks are consistently stronger than vanilla \pgd, reducing defended accuracy on both ImageNet\textnormal{-}D and VQA-v2.
However, PDA does not collapse under any \eot variant and retains a clear robustness margin over the no-defense baseline, indicating that its gains are not merely due to attacker mismatch against a stochastic pipeline.
Overall, agreement-based aggregation over diversified textual views remains effective even when the attacker explicitly targets the view distribution induced by PDA.}

\begin{table}[t]
\centering
\renewcommand{\arraystretch}{1.05}
\setlength{\tabcolsep}{8pt}
\caption{Robustness under more challenging white-box attacks on ImageNet\textnormal{-}D and VQA-v2.}
\label{tab:stronger_whitebox}
\begin{tabular}{lcccc}
\toprule
\multirow{2}{*}{Setting}
& \multicolumn{2}{c}{\textbf{No Defense}}
& \multicolumn{2}{c}{\textbf{PDA}} \\
\cmidrule(lr){2-3}\cmidrule(lr){4-5}
& ImageNet-D & VQA-v2 & ImageNet-D & VQA-v2 \\
\midrule
\multicolumn{5}{l}{\textit{(a) Larger $\epsilon$ under \pgd}} \\
$2/255$ & 0.25 & 0.45 & 0.57 & 0.75 \\
$4/255$ & 0.22 & 0.42 & 0.56 & 0.74 \\
$6/255$ & 0.19 & 0.40 & 0.53 & 0.74 \\
$8/255$ & 0.17 & 0.39 & 0.54 & 0.72 \\
\midrule
\multicolumn{5}{l}{\textit{(b) Adaptive \eot attacks}} \\
PGD    & 0.25 & 0.45 & 0.57 & 0.75 \\
EOT-P  & 0.20 & 0.41 & 0.53 & 0.72 \\
EOT-D  & 0.23 & 0.40 & 0.55 & 0.70 \\
EOT-PD & 0.21 & 0.42 & 0.54 & 0.74 \\
\bottomrule
\end{tabular}
\vspace{-4mm}
\end{table}

\subsection{Robustness against Black-box Attacks}

\noindent \textbf{General Black-box Attacks.} \rev{Besides white-box robustness, black-box robustness is also critical in practice because many deployed LVLMs are accessed through closed APIs where gradients and model parameters are unavailable. This is particularly relevant to PDA: since our defense is designed to be a test-time wrapper that only interacts with the model through its textual interface, its usefulness should not hinge on privileged white-box access. We therefore complement the white-box evaluation with representative black-box attacks to assess whether PDA’s gains persist under realistic constraints.}

\begin{wraptable}{r}{0.48\linewidth}
\centering
\vspace{-16pt}
\renewcommand{\arraystretch}{1.1} 
\setlength{\tabcolsep}{10pt} 
\caption{Black-box attacks on ImageNet\textnormal{-}D and VQA-v2. Each entry is \textbf{No Defense / PDA} robust accuracy.}
\label{tab:blackbox}
\begin{tabular}{lcc}
\toprule
Attack & ImageNet-D & VQA-v2 \\
\midrule
Transfer-based~\cite{Cui2024RobustnessCVPR} & 0.58 / 0.62 & 0.77 / 0.81 \\
AttackVLM~\cite{AttackVLM} & 0.56 / 0.61 & 0.75 / 0.79 \\
AnyAttack~\cite{Zhang2025AnyAttackCVPR} & 0.61 / 0.64 & 0.79 / 0.81 \\
\bottomrule
\end{tabular}
\vspace{-8pt}
\end{wraptable}

\rev{Following prior LVLM robustness studies, we consider three black-box settings that cover common attacker capabilities (Table~\ref{tab:blackbox}). The first is a transfer-based black-box setting~\cite{Cui2024RobustnessCVPR}, where adversarial perturbations are generated on a surrogate visual encoder and then evaluated on the target LVLM without access to its gradients. We additionally include AttackVLM~\cite{AttackVLM} and AnyAttack~\cite{Zhang2025AnyAttackCVPR}, two representative black-box attack frameworks that have been widely used to benchmark practical robustness of multimodal systems. For all three attacks, we use the default configurations from the original papers and keep the datasets, evaluation metrics, and prompting templates identical to our white-box experiments; each entry reports the performance of the same victim model with \emph{No Defense} and with \emph{PDA} enabled.}

\rev{As shown in Table~\ref{tab:blackbox}, PDA consistently improves robustness across all three black-box attacks on both ImageNet\textnormal{-}D and VQA-v2. We also observe that the absolute accuracies under black-box attacks are higher than those under strong white-box \pgd, because black-box attacks, including transfer-based and finite-query settings, are generally weaker than strong white-box attacks. Importantly, the robustness gains from PDA remain stable across attack families, suggesting that PDA does not rely on fragile white-box assumptions. This behavior aligns with the core mechanism of PDA: paraphrase and decomposition generate multiple semantically equivalent views of the same input, and agreement-based aggregation suppresses sporadic failures that do not persist across views, making the defense effective even when the adversary can only interact with the model through black-box feedback.}

\begin{wraptable}{r}{0.5\linewidth} 
\centering
\vspace{-16pt}
\renewcommand{\arraystretch}{1.1}
\setlength{\tabcolsep}{8pt} 
\caption{Closed-source LVLMs under transfer attacks.
Each entry is \textbf{ImageNet\textnormal{-}D / VQA-v2} robust accuracy.}
\label{tab:closed_ext}
\begin{tabular}{lccc}
\toprule
 & GPT-5 & Claude & Gemini \\
\midrule
No Defense & 0.67 / 0.81 & 0.61 / 0.82 & 0.67 / 0.76 \\
PDA & 0.76 / 0.90 & 0.70 / 0.88 & 0.72 / 0.86 \\
\bottomrule
\end{tabular}
\vspace{-16pt}
\end{wraptable}

\noindent \textbf{Closed-source LVLMs under Transfer Attacks.}
\rev{To reflect realistic deployment where the victim LVLM is only accessible through commercial APIs, we further evaluate PDA on closed-source models using a transfer-attack protocol (Table~\ref{tab:closed_ext}).
Specifically, we craft adversarial images with white-box \pgd on LLaVA and directly transfer these adversarial images to closed-source APIs for evaluation, keeping the input images and prompts fixed and measuring task performance in the same way as in our open-source experiments.
This setting is a standard and reproducible way to probe robustness of API-based systems, since it avoids relying on unrestricted query budgets or API-specific gradient estimators while still capturing adversarial vulnerability under realistic access constraints.}

\rev{Table~\ref{tab:closed_ext} shows that PDA consistently improves robustness across all tested closed-source LVLMs on both ImageNet\textnormal{-}D and VQA-v2.
The gains are observed for GPT-5, Claude, and Gemini, indicating that PDA can be applied as a black-box wrapper even when the underlying model family is proprietary and its internals are unknown.
These results complement our query-based black-box evaluation: together, they suggest that PDA’s robustness benefits persist in practical deployment scenarios, including both iterative black-box attack settings and transfer-based attacks against commercial LVLM APIs.}

\subsection{Comparison with Test-time Baselines.}
\label{subsec:more_defenses}

\noindent \textbf{Image-side Defenses.}
\rev{Beyond text-side randomized smoothing and training-time robustification baselines (Table~\ref{tab:main}), we additionally compare PDA with lightweight \emph{image-side} test-time defenses that are widely used as practical pre-processing modules for vision models.
Specifically, we consider JPEG compression and random augmentation (RA), which aim to attenuate pixel-level perturbations by either removing high-frequency artifacts (JPEG) or injecting benign input variation (RA) that can partially wash out adversarial patterns.
These methods are attractive in deployment because they are model-agnostic and require no changes to model weights or prompting, but they also operate purely on the visual stream and do not exploit the textual modality that LVLMs naturally provide.}

\rev{We evaluate all defenses under the same white-box $\ell_\infty$ \pgd threat model ($\epsilon{=}2/255$) with identical datasets, metrics, and prompting templates.
As shown in Table~\ref{tab:baselines_ext}(a), both JPEG and RA improve robustness over the no-defense baseline, but they remain clearly behind PDA on both ImageNet\textnormal{-}D and VQA-v2.
More importantly, combining them with PDA yields further gains (e.g., JPEG+PDA and RA+PDA), suggesting that PDA is largely \emph{orthogonal} to image-side defenses: pre-processing reduces the strength of pixel-level corruption, while PDA diversifies the textual views and aggregates answers to suppress sporadic failures that do not persist across views.
This complementarity makes PDA a practical plug-in that can be stacked with standard pre-processing without additional training.}

\noindent \textbf{Reasoning Baselines.}
\rev{We also compare PDA with \emph{reasoning-based} test-time baselines that attempt to recover correct answers by eliciting more structured inference from the victim LVLM, rather than modifying the input image.
We consider CoT prompting (LLaVA-CoT) and self-consistency voting, both of which are commonly used to improve reliability in clean settings by encouraging explicit intermediate reasoning and by aggregating multiple sampled outputs.
In the adversarial setting, however, these methods operate on the same corrupted visual evidence and thus primarily target errors \emph{after} the perception stage.}

\rev{Table~\ref{tab:baselines_ext}(b) shows that these reasoning baselines provide limited robustness gains and significantly underperform PDA.
This gap highlights a key intuition in adversarial VLM robustness: when the image perturbation biases visual perception, generating longer rationales or sampling multiple reasoning traces cannot reliably restore the missing or distorted visual cues, and may even amplify spurious evidence.
In contrast, PDA actively diversifies the \emph{queries} to probe stable visual evidence from multiple angles and then aggregates for agreement, making it more effective for correcting perception-level corruption.
Overall, these results position PDA as a stronger test-time defense than purely reasoning-based strategies under the same threat model.}

\begin{table}[h]
\centering
\renewcommand{\arraystretch}{1.05}
\setlength{\tabcolsep}{18pt}
\caption{Comparisons with additional test-time baselines.}
\label{tab:baselines_ext}
\begin{tabular}{lcc}
\toprule
Method & ImageNet-D & VQA-v2 \\
\midrule
No Defense & 0.25 & 0.45 \\
PDA & 0.57 (+0.32) & 0.75 (+0.30) \\
\midrule
\multicolumn{3}{l}{\textit{(a) Image-side defenses}} \\
JPEG & 0.51 (+0.26) & 0.68 (+0.23) \\
JPEG + PDA & 0.61 (+0.36) & 0.77 (+0.32) \\
RA & 0.55 (+0.30) & 0.70 (+0.25) \\
RA + PDA & \textbf{0.67 (+0.42)} & \textbf{0.80 (+0.35)} \\
\midrule
\multicolumn{3}{l}{\textit{(b) Reasoning baselines}} \\
LLaVA-CoT & 0.53 (+0.28) & 0.51 (+0.06) \\
Self-consistency & 0.32 (+0.07) & 0.49 (+0.04) \\
\bottomrule
\end{tabular}
\vspace{-4mm}
\end{table}

\subsection{Ablation Studies and Design Choices}
\label{subsec:ablations}

\noindent \textbf{Variants across Tasks and Architectures.}
To evaluate the effectiveness of our proposed variants, we conduct controlled comparisons across different tasks and backbone architectures under a consistent threat model and decoding protocol.
Table~\ref{tab:vqa_ablation} compares the \emph{budgeted} PDA variants across VQA-v2 and ImageNet\textnormal{-}D.
All PDA variants outperform the no-defense baseline, though the optimal choice depends on the specific task--backbone pair.
On ImageNet\textnormal{-}D, \emph{PDA--PV} generally performs best: in zero-shot CLIP-style classification, class texts are often closely spaced, making it more reliable to verify a few competing labels than to generate long, detail-heavy rationales, which can introduce non-visual hallucinations and sway the decision.
Conversely, in VQA, the answer space is compact while the evidential path is heterogeneous; here, revealing more verifiable cues through richer decomposition (e.g., \emph{PDA--RJV}) is generally beneficial.

\begin{wraptable}{r}{0.55\linewidth} 
\centering
\setlength{\tabcolsep}{4pt}
\renewcommand{\arraystretch}{1.1}
\vspace{-16pt}
\caption{\textbf{Ablation of budgeted PDA variants across backbone architectures.}
We report robustness (\%) of each victim VLM under the same adversarial threat model, with values shown as ``VQA-v2 / ImageNet-D.''}
\label{tab:vqa_ablation}
\resizebox{\linewidth}{!}{%
\begin{tabular}{lcccc}
\toprule
Method
& \textbf{DeepSeek-VL}
& \textbf{InternVL3}
& \textbf{LLaVA-1.5-7B}
& \textbf{Ovis2} \\
\midrule
No Defense
& 59.0 / 46.4
& 69.6 / 25.8
& 47.2 / 26.8
& 88.0 / 39.6 \\
PDA--RJV
& 63.8 / 53.4
& \textbf{76.0} / \textbf{39.4}
& \textbf{65.2} / 49.0
& 87.4 / 53.2 \\
PDA--RDA
& \textbf{63.4} / 51.8
& 72.4 / 34.0
& 61.2 / 45.8
& \textbf{89.8} / 48.4 \\
PDA--PV
& 61.6 / \textbf{70.2}
& 72.2 / 31.4
& 58.4 / \textbf{57.0}
& 89.2 / \textbf{56.4} \\
\bottomrule
\end{tabular}
}
\vspace{-8pt}
\end{wraptable}
Although  PDA-Full achieves the strongest VQA robustness on DeepSeek-VL and LLaVA-1.5-7B, showing that more exhaustive multi-view reasoning can further improve robustness on some backbones.
However, the budgeted variants remain preferable overall: PDA--RJV attains the best VQA accuracy on InternVL3, PDA--RDA is slightly superior on Ovis2, and all three budgeted variants incur substantially lower inference cost.
On ImageNet-D, the per-sample runtime of each method is as follows: the model without defense takes 4.5 seconds per sample.
Among the PDA variants, PDA-Full is the most expensive at 15.5 seconds, followed by PDA-RJV at 10.5 seconds, PDA-RDA at 6.0 seconds, and PDA-PV, which is the fastest, at 5.5 seconds.
All measurements are obtained using LLaVA-1.5-13B as the victim model and DeepSeek-Chat as the aggregator on an RTX A6000 GPU.


\begin{wraptable}{r}{0.5\linewidth} 
\centering
\vspace{-16pt}
\setlength{\tabcolsep}{4pt}
\renewcommand{\arraystretch}{1.05}
\caption{Effect of the number of paraphrases $K$ on ADV accuracy for PDA--RJV and PDA--RDA on ImageNet-D.}
\label{tab:k_sweep_imagenetd}
\resizebox{\linewidth}{!}{%
\begin{tabular}{@{}llcccc@{}}
\toprule
\textbf{Variant} & \textbf{Backbone} & \textbf{$K=3$} & \textbf{$K=5$} & \textbf{$K=7$} & \textbf{$K=9$} \\
\midrule
\multirow{4}{*}{PDA--RJV}
 & LLaVA-1.5-7B & 49.2 & \textbf{53.2} & 53.0 & 52.2 \\
 & InternVL3    & \textbf{40.8} & 39.2          & 29.2 & 39.0 \\
 & DeepSeek-VL  & 53.4 & \textbf{53.4} & 53.2 & 52.4 \\
 & Ovis-2       & 51.4 & \textbf{53.2} & 52.2 & 52.4 \\
\midrule
\multirow{4}{*}{PDA--RDA}
 & LLaVA-1.5-7B & \textbf{47.0} & 43.7 & 43.9 & 43.5 \\
 & InternVL3    & \textbf{34.2} & 29.3 & 30.3 & 29.3 \\
 & DeepSeek-VL  & 51.8 & 51.8 & 51.2 & \textbf{52.2} \\
 & Ovis-2       & 47.0 & \textbf{48.4} & 47.6 & \textbf{48.4} \\
\bottomrule
\end{tabular}
}
\vspace{-12pt}
\end{wraptable}

\begin{figure}
  \centering
  \vspace{-10pt}
  \includegraphics[width=0.6\linewidth]{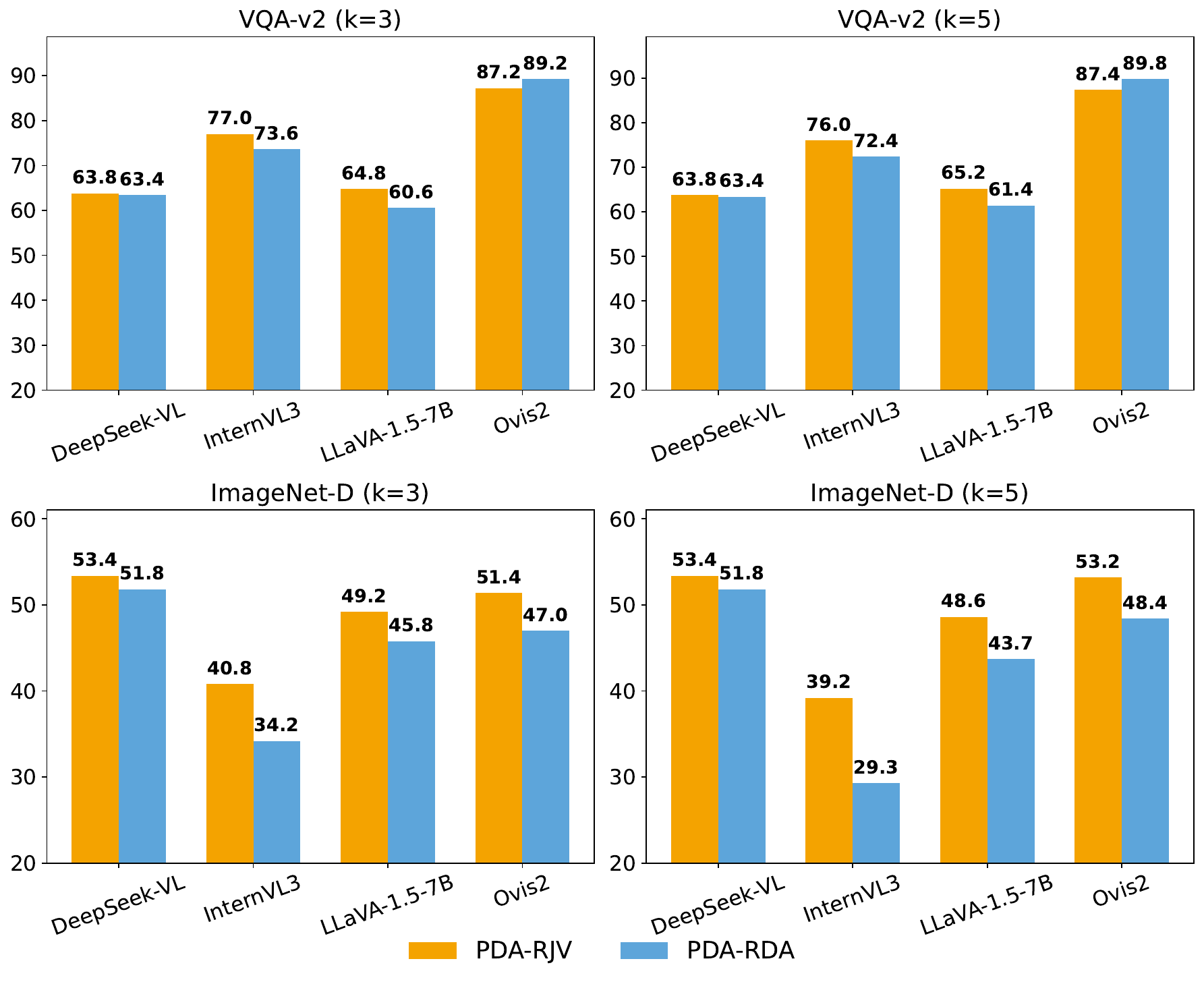}
  \caption{Overall performance of PDA variants on VQA-v2 and ImageNet-D under different paraphrase counts ($K{=}3$ and $K{=}5$).}
  \label{fig:pda_k}
  \vspace{-10pt}
\end{figure}

\noindent \textbf{Effect of the Number of Paraphrases}
We vary the paraphrase count $K$ to study the robustness--efficiency trade-off under a fixed agent and threat model.
Figure~\ref{fig:pda_k} compares the representative settings $K\in\{3,5\}$ across tasks and backbones.
The gap between $K{=}3$ and $K{=}5$ is generally small, and neither choice uniformly dominates across all settings.
To further probe whether larger paraphrase widths help, Table~\ref{tab:k_sweep_imagenetd} extends the sweep to $K\in\{3,5,7,9\}$ for PDA--RJV and PDA--RDA on ImageNet-D.

Overall, robustness tends to saturate quickly once $K$ reaches a moderate range.
Across backbones, most differences remain within a few percentage points, and there is no consistent benefit from very large $K$: the best or near-best results typically occur at $K{=}3$ or $K{=}5$, while $K{=}7$ and $K{=}9$ often bring no further gains and occasionally degrade performance.
We attribute this to two factors:
(i) longer prompts from additional paraphrases increase context length and dispersion, which can hinder retrieval of the truly relevant cues and dilute aggregation; and
(ii) marginal paraphrases are noisier and may invite non-visual hallucination.
In practice, these results suggest that performance saturates early; we use $K{=}5$ as a stable default in the main experiments, while $K{=}3$ remains an attractive lower-cost alternative with very similar robustness.

\begin{table}[h]
\centering
\footnotesize
\setlength{\tabcolsep}{5pt}
\renewcommand{\arraystretch}{1.12}
\caption{\textbf{Effect of LLM agent within the PDA pipeline} using $K{=}5$.
Each cell reports \textbf{VQA-v2 / ImageNet\textnormal{-}D} robust accuracy (\%) for the same victim VLM under different LLM agents.}
\label{tab:agents_k5}
\resizebox{0.67\columnwidth}{!}{%
\begin{tabular}{lcccc}
\toprule
Agent
& \textbf{DeepSeek-VL}
& \textbf{InternVL3}
& \textbf{LLaVA-1.5-7B}
& \textbf{Ovis2} \\
\midrule
DeepSeek
& \textbf{63.4} / 54.4
& 70.8 / 35.7
& 60.8 / 49.2
& \textbf{90.0} / \textbf{49.2} \\
GPT
& \textbf{63.4} / 51.8
& \textbf{72.4} / 29.3
& \textbf{61.4} / 43.7
& 89.8 / 48.4 \\
LLaMA
& 58.0 / \textbf{54.6}
& 58.4 / \textbf{36.1}
& 58.0 / \textbf{50.4}
& 61.2 / 41.2 \\
Qwen
& 63.2 / 50.6
& 70.2 / 34.8
& \textbf{61.4} / 47.1
& 89.2 / 47.2 \\
\bottomrule
\end{tabular}
}
\vspace{-2mm}
\end{table}

\noindent \textbf{Effect of LLM Agent.}
Using the same PDA pipeline, we compare four agents: a commercial small model (\textit{GPT-4.1 mini}), a commercial open API model (\textit{DeepSeek-Chat}), and two locally deployable instruct models (\textit{Llama-3.1-8B-Instruct}, \textit{Qwen2.5-14B-Instruct}), as shown in Table~\ref{tab:agents_k5}.
On VQA-v2, stronger instruction followers tend to help decomposition: GPT is best or tied on three backbones and ties DeepSeek-VL.
On ImageNet\textnormal{-}D, where pairwise checks dominate, capability gaps narrow and the lightweight local models are competitive or best.
Practically, this suggests pairing \emph{decomposition-heavy} settings with a stronger agent and \emph{verification-heavy} settings with a compact agent for better accuracy--performance trade-offs.

%% file: sec/tables/overall.tex
\begin{table*}[t]
\centering
\footnotesize
\setlength{\tabcolsep}{3pt}
\renewcommand{\arraystretch}{1.05}
\caption{\textbf{Overall comparison on three tasks.} Results for VQA-v2 (\emph{VQA Accuracy}), ImageNet\textnormal{-}D (\emph{Top-1 Accuracy}), and MS~COCO (\emph{CLIPScore}). Rows list defenses applied to each victim model (LLaVA-1.5-7B/13B), including SmoothLLM (Swap/Insert/Patch), SemanticSmooth (Paraphrase/Summarise), TeCoA, FARE, and our \textbf{PDA}. \emph{Clean} denotes performance on unperturbed inputs; \emph{ADV} denotes performance on adversarially perturbed inputs under the same evaluation protocol.}

\begin{tabular*}{\textwidth}{@{\extracolsep{\fill}}lcccccccccccc@{}}
\toprule
\multirow{3}{*}{Defense Method} 
& \multicolumn{6}{c}{\textbf{LLaVA-1.5-7B}} 
& \multicolumn{6}{c}{\textbf{LLaVA-1.5-13B}} \\
\cmidrule(lr){2-7} \cmidrule(lr){8-13}
& \multicolumn{2}{c}{VQA-v2} 
& \multicolumn{2}{c}{ImageNet-D} 
& \multicolumn{2}{c}{COCO}
& \multicolumn{2}{c}{VQA-v2} 
& \multicolumn{2}{c}{ImageNet-D} 
& \multicolumn{2}{c}{COCO} \\
\cmidrule(lr){2-3} \cmidrule(lr){4-5} \cmidrule(lr){6-7}
\cmidrule(lr){8-9} \cmidrule(lr){10-11} \cmidrule(lr){12-13}
& Clean & ADV & Clean & ADV & Clean & ADV
& Clean & ADV & Clean & ADV & Clean & ADV \\
\midrule
No Defense                 
& 0.862 & \cellcolor{gray!12}0.445 & 0.625 & \cellcolor{gray!12}0.248 & 0.858 & \cellcolor{gray!12}0.753
& \textbf{0.886} & \cellcolor{gray!12}0.593 & 0.646 & \cellcolor{gray!12}0.320 & \textbf{0.865} & \cellcolor{gray!12}0.757 \\
SmoothLLM-Swap             
& 0.866 & \cellcolor{gray!12}0.453 & 0.610 & \cellcolor{gray!12}0.260 & 0.849 & \cellcolor{gray!12}0.744
& 0.881 & \cellcolor{gray!12}0.596 & 0.630 & \cellcolor{gray!12}0.327 & 0.849 & \cellcolor{gray!12}0.756 \\
SmoothLLM-Insert           
& 0.870 & \cellcolor{gray!12}0.449 & 0.596 & \cellcolor{gray!12}0.254 & 0.854 & \cellcolor{gray!12}0.746
& 0.884 & \cellcolor{gray!12}0.587 & 0.605 & \cellcolor{gray!12}0.330 & 0.855 & \cellcolor{gray!12}0.743 \\
SmoothLLM-Patch            
& 0.864 & \cellcolor{gray!12}0.454 & 0.632 & \cellcolor{gray!12}0.261 & 0.843 & \cellcolor{gray!12}0.741
& 0.877 & \cellcolor{gray!12}0.599 & 0.611 & \cellcolor{gray!12}0.314 & 0.851 & \cellcolor{gray!12}0.751 \\
SemanticSmooth-Paraphrase  
& 0.850 & \cellcolor{gray!12}0.452 & 0.638 & \cellcolor{gray!12}0.276 & 0.850 & \cellcolor{gray!12}0.748
& 0.884 & \cellcolor{gray!12}0.608 & 0.657 & \cellcolor{gray!12}0.358 & 0.855 & \cellcolor{gray!12}0.751 \\
SemanticSmooth-Summarise   
& 0.846 & \cellcolor{gray!12}0.456 & 0.610 & \cellcolor{gray!12}0.289 & 0.851 & \cellcolor{gray!12}0.750
& 0.870 & \cellcolor{gray!12}0.613 & 0.649 & \cellcolor{gray!12}0.379 & 0.861 & \cellcolor{gray!12}0.754 \\
TeCoA                      
& 0.787 & \cellcolor{gray!12}0.689 & 0.613 & \cellcolor{gray!12}0.569 & 0.811 & \cellcolor{gray!12}0.783
& 0.830 & \cellcolor{gray!12}0.672 & 0.697 & \cellcolor{gray!12}0.611 & 0.823 & \cellcolor{gray!12}0.791 \\
FARE                       
& 0.828 & \cellcolor{gray!12}0.703 & 0.640 & \cellcolor{gray!12}0.511 & 0.835 & \cellcolor{gray!12}\textbf{0.806}
& 0.838 & \cellcolor{gray!12}0.729 & 0.695 & \cellcolor{gray!12}\textbf{0.625} & 0.833 & \cellcolor{gray!12}\textbf{0.805} \\
\textbf{PDA}               
& \textbf{0.871} & \cellcolor{gray!12}\textbf{0.754} & \textbf{0.758} & \cellcolor{gray!12}\textbf{0.570} & \textbf{0.860} & \cellcolor{gray!12}0.774
& 0.879 & \cellcolor{gray!12}\textbf{0.787} & \textbf{0.798} & \cellcolor{gray!12}0.566 & 0.862 & \cellcolor{gray!12}0.778 \\
\bottomrule
\end{tabular*}

\label{tab:main}
\vspace{-11pt}
\end{table*}

%% file: sec/5_conclusion.tex
\section{Conclusion}
We presented \emph{PDA}, a training-free, text-side defense that stabilizes VLM predictions under adversarial image perturbations by \emph{paraphrasing} the original query into diverse yet semantically equivalent prompts, \emph{decomposing} them into atomic questions, and \emph{aggregating} the resulting answers with agreement-aware voting at inference. To validate practicality and generality, we evaluated propossed framework across different tasks and multiple backbones, observing consistent robustness gains with minimal clean degradation. And on some datasets like ImageNet\textnormal{-}D, PDA even improved clean accuracy due reducing spurious predictions by cross-checked evidence. For efficiency, we further explored compressed variants that retain most of the robustness while lowering latency, yielding task-dependent recipes. We hope this perspective on training-free, inference-time robustness will spur research on principled certifications and stronger attack settings, ultimately advancing the safety of VLMs in real-world use.

%% file: sec/X_suppl.tex
\clearpage



\appendix

\section{Overview}
\label{sec:supp_overview}

This supplementary material provides additional details and results for our PDA framework beyond what is reported in the main paper.
It is organized into two parts:
(i) full prompt specifications for all text-only LLM agents used in PDA, including semantic and logical paraphrasing, decomposition, and caption verification/aggregation and
(ii) additional qualitative case studies on VQA-v2, ImageNet-D, and COCO captioning that illustrate typical failure modes and how PDA corrects them.

\section{Prompts for LLM Agents}
\label{sec:supp_prompts}

In this section, we document the exact prompts used to interact with the text-only LLMs in PDA.  For reproducibility, we report system, user, and (when applicable) few-shot example templates. All prompts follow the same high-level design principles as discussed in Section~\ref{sec:method} of the main paper, but are presented here in full for completeness.

\subsection{Prompts for Paraphrase }
\label{subsec:prompt_paraphrase}

The paraphrase agent is prompted as a ``Paraphrase Engine’’ that performs only local synonym or short-phrase substitutions, so that we gain textual diversity without changing the underlying decision problem.  The strict rules fix negation, quantities, and proper nouns and preserve the question type, which prevents the LLM from implicitly altering the task (e.g., by re-framing or summarising the query).  

We first provide the system role description, followed by the user instruction and an illustrative example. The final prompt used in our experiments can be pasted here as a verbatim code block. The adjustable parameters (\texttt{change\_intensity}, \texttt{num\_candidates}) expose a simple knob for controlling diversity, while the JSON-only output constraint removes free-form reasoning text and makes it easy to parse and filter the $N$ candidates.

\noindent\textbf{(a) System prompt (Semantic paraphrase).}

\begin{promptbox}
You are a Paraphrase Engine.

Task: Rewrite the input sentence using synonym / short-phrase substitutions
while preserving the same meaning. Keep the structure and length roughly similar.

Strict rules:
- Do not add or remove facts.
- Preserve negation, comparison, quantities, times, names, and proper nouns
  (keep their casing).
- Keep the sentence type unchanged (questions stay questions, statements stay statements).
- Prefer part-of-speech aligned swaps (noun<->noun, verb<->verb, etc.).
  If no good swap exists, keep the original token.
- No explanations, no reasoning - output only the rewritten sentence(s).
- Match the language of the input (English in -> English out; Chinese in -> Chinese out).

Adjustable parameters:
- change_intensity: low | medium | high
- num_candidates: integer K

Output format:
Return a single JSON object: {"candidates": [...]}
- The array length must equal K.
- Each element is one paraphrased sentence (string).
- No extra text before/after the JSON.
\end{promptbox}

\noindent\textbf{(b) User prompt (Semantic paraphrase).}

\begin{promptbox}
Now start:
- change_intensity: medium
- num_candidates: 4
- input_sentence: <ORIGINAL SENTENCE HERE>
\end{promptbox}

\noindent\textbf{(c) System prompt (Logical paraphrase for PDA-PV).}

\begin{promptbox}
Request:
Input:
(Example) Which object is more visually prominent in the image?
Choose from the following list: jeans, t shirt.

Generation steps:
To generate logically equivalent contrasting questions, consider the
following attributes:
- Material: different types of materials the objects are made from
  (e.g., denim vs cotton).
- Shape: the overall shape or outline of the object
  (e.g., pant shape vs shirt shape).
- Color: the color or color pattern of the object
  (e.g., blue jeans vs white t-shirt).
- Functional components: elements related to functionality
  (e.g., fitted pants vs fitted shirt).
- State of use: how the object is used or worn
  (e.g., worn on lower body vs worn on upper body).
- Position/area: the area of the body or region in focus
  (e.g., leg area vs chest area).
- Pairing: how the objects are paired with other items
  (e.g., shoes with pants vs accessories with shirt).

Logic equivalence:
The new questions must be logically equivalent to the original question.
If the first option should be selected in the original question, the
first option in every generated question should also be the logically
correct one. This preserves the decision boundary while changing the
surface form.

Output requirements:
- Generate TEN logically equivalent multiple-choice questions.
- Each question should use only the object names in the options, without
  extra explanations in parentheses.
- Return a single JSON object with the structure:

{
  "original_question":
    "Which object is more visually prominent in the image? \
Choose from the following list: (A) jeans, (B) t shirt.",
  "generated_questions": [
    {
      "question":
        "Which object is more visually prominent in the image? \
Choose from the following list: (A) lower body wear, (B) upper body wear.",
      "options": ["lower body wear", "upper body wear"]
    },
    {
      "question":
        "Which object is more visually prominent in the image? \
Choose from the following list: (A) pants, (B) shirt.",
      "options": ["pants", "shirt"]
    },
    {
      "question":
        "Which object is more visually prominent in the image? \
Choose from the following list: (A) leg area, (B) chest area.",
      "options": ["leg area", "chest area"]
    },
    {
      "question":
        "Which object is more visually prominent in the image? \
Choose from the following list: (A) denim pants, (B) cotton shirt.",
      "options": ["denim pants", "cotton shirt"]
    },
    {
      "question":
        "Which object is more visually prominent in the image? \
Choose from the following list: (A) clothing for legs, (B) clothing for torso.",
      "options": ["clothing for legs", "clothing for torso"]
    }
    ... 5 more question objects ...
  ]
}
\end{promptbox}

For PDA-PV, aggregation is reduced to simple voting, so we need the paraphrase layer itself to explore richer but still \emph{logically equivalent} variants of the original multiple-choice query.  The above prompt asks the LLM to rewrite the question by contrasting different attributes (material, position, body area, etc.) while preserving the option-wise decision boundary: whenever option A is correct originally, option A must also be correct in every generated question.  Constraining
the output to a JSON list of questions and option strings makes it easy to parse these logical paraphrases and reuse them directly as VLM queries in PDA-PV.

\subsection{Prompts for Decomposition}
\label{subsec:prompt_decomposition}

We now describe the prompts used to decompose a main image question or caption into sub-questions. For generic VQA-style tasks (VQA-v2, ImageNet-D), the LLM directly designs factual sub-questions for the given query; for captioning, we first extract structured atomic claims from the short caption and then generate verification sub-questions conditioned on these claims.

\noindent\textbf{(a) System prompt (Decomposition for VQA/ImageNet-D).}

\begin{promptbox}
You are a visual reasoning expert. Generate 3-5 factual sub-questions to answer an image-based question WITHOUT seeing the image. Follow these rules:

Security & Robustness Rules:

Use descriptive phrases instead of specific nouns from the original question (e.g., "long-necked animal" instead of "giraffe")
Design redundancy: Key attributes should be cross-verified by multiple questions
Include error-checking questions to catch VLM inconsistencies
Use negative verification questions where applicable
Make questions independent to isolate potential errors

Output Format:
{
  "original_question": "[exact input]",
  "sub_questions": [
    {
      "id": 1,
      "question": "[simple complete question]",
      "answer_type": "yes_no/choice/phrase",
      "options": "[if choice]"
    },
    ...
  ],
  "answer_logic": "[Concise 1-sentence reasoning]"
}

Example:
Input: "Is this rice noodle soup?"
Output:
{
  "original_question": "Is this rice noodle soup?",
  "sub_questions": [
    {
      "id": 1,
      "question": "Are thin noodle strands visible?",
      "answer_type": "yes_no"
    },
    {
      "id": 2,
      "question": "What noodle texture appears?",
      "answer_type": "choice",
      "options": "translucent/opaque/other"
    },
    {
      "id": 3,
      "question": "Is liquid broth present?",
      "answer_type": "yes_no"
    },
    {
      "id": 4,
      "question": "What color are the noodles?",
      "answer_type": "phrase"
    }
  ],
  "answer_logic": "If Q1=yes and Q2=translucent and Q3=yes -> likely rice noodle soup; otherwise not."
}

Now process:
\end{promptbox}

For generic VQA and ImageNet-D, this decomposition prompt treats the LLM as a visual reasoning expert that proposes 3--5 factual, robustness-aware sub-questions plus a concise \texttt{answer\_logic} sentence, encouraging descriptive rephrasings, redundancy, and explicit decision rules so that the downstream VLM+LLM pipeline can cross-verify key attributes instead of relying on a single potentially fragile query.

\noindent\textbf{(b) System prompt (Claim extraction for Caption Verification).}

\begin{promptbox}
You extract atomic visual claims from a short COCO-style caption.

You MUST return a single JSON object with EXACTLY these keys:

- "subject_head": a short noun for the main subject, such as
  "woman", "man", "dog", "cat", "car", "bicycle", "kitchen",
  "bathroom", or "unknown" if you are not sure.
- "subject_count": one of ["one", "two", "many", "unknown"].
- "key_object": a short noun for ONE most important secondary object,
  such as "umbrella", "bench", "truck", "clock". If there is no clear
  secondary object, use "none" or "unknown".
- "relation": a short phrase for the relation between subject and
  key_object, such as "holding", "sitting on", "next to",
  "leaning against", "in front of", or "none"/"unknown" if unclear.
- "scene": one or two words for coarse scene type, such as
  "street", "kitchen", "bathroom", "bedroom", "field", or "unknown"
  if the scene type is not stated.

Rules:
- Base everything ONLY on the given caption.
- Do NOT invent objects or places not mentioned in the caption.
- If something is not stated, use "unknown" (or "none" where appropriate).
- Be concise; prefer single words when possible.

Output format (IMPORTANT):
- Return ONLY the JSON object, no extra text.
\end{promptbox}

\noindent\textbf{User prompt template (Claim extraction).}

\begin{promptbox}
Short caption:
---
{SHORT_CAPTION}
---

Extract the JSON object with keys:
"subject_head", "subject_count", "key_object", "relation", "scene".
\end{promptbox}

For caption-based tasks, we first run this claim-extraction stage to convert the short COCO-style caption into a structured set of atomic claims (main subject, count, key object, relation, scene); these claims are then passed as part of the input to the verification-oriented decomposition prompt below, which allows us to design targeted sub-questions that directly stress-test the caption’s core semantics.

\noindent\textbf{(c) System prompt (Decomposition for Caption Verification).}

\begin{promptbox}
You design verification sub-questions for an image, focusing on finding possible conflicts with a short COCO-style caption.

Inputs:
- Parsed short-caption claims (subject_head, subject_count, key_object, relation, scene).
- One detailed caption of the image (high-recall, possibly noisy).

Goal:
Write 3-5 short questions that will be asked to a vision-language model that CAN see the image, to verify the short-caption's core claims:
- main subject identity (from subject_head),
- subject count (one/two/many),
- key secondary object (key_object),
- the relation between the subject and the key object (relation),
- a coarse scene type (scene) if the caption clearly asserts one.

Design them so that:
- If the short caption is wrong about these aspects, the answers are likely to contain "no" or "unclear".
- At least one question explicitly checks the number of main subjects.
- At least one question explicitly checks the presence and identity of the key_object (if it is not "none"/"unknown").
- At least one question explicitly checks the relation between subject and key_object (if relation is not "none"/"unknown").
- Optionally, one question checks the coarse scene type when the caption asserts a clear scene (e.g., "kitchen", "bathroom", "street").

COUNT-FOCUSED DESIGN (very important):
- Let claims.subject_count belong {"one", "two", "many", "unknown"}.
- If claims.subject_count is "one", "two", or "many", you MUST create at least two different questions about the number of main subjects:
  1) One generic count question whose last sentence is:
     "Answer exactly: 'one', 'two', 'many', 'none', or 'unclear'."
  2) One yes/no question that directly tests the asserted count, for example:
     "Does the image show exactly two women as the main subjects?"
     "Is there exactly one dog as the main subject?"
     and this yes/no question must end with:
     "Answer exactly: 'yes', 'no', or 'unclear'."
- If claims.subject_count is "unknown", you may include only the generic count question (or even skip it) and focus more on subject identity and key object.

APPEARANCE CHECKS FOR MAIN OBJECT:
- Use claims.subject_head as the main subject type.
- If claims.subject_head is very generic (e.g., "person", "people", "man", "woman", "child", "animal", "vehicle", "object", "thing", "room"), you may keep appearance questions very short or skip one of them.
- If claims.subject_head is more specific (e.g., "toilet", "bathtub", "telephone", "sink", "bicycle", "bench", "stove", "laptop", "jeep"), you MUST include appearance checks so that the model is forced to describe what the main object looks like.

In particular, when subject_head is specific, you MUST add:
- One question asking the model to briefly describe the main subject's visual appearance (color, material, rough shape), for example:
  "Describe the main subject's appearance (color and material) in at most five words. Answer in up to five words."
- Optionally, one yes/no question asking whether the main subject looks like a typical <subject_head>, for example:
  "Does the main subject look like a typical toilet fixture you would see in a bathroom? Answer exactly: 'yes', 'no', or 'unclear'."
- These appearance questions are intended to catch confusions such as mistaking a toilet for an old telephone, or a sink for a bathtub.

Constraints:
- Most questions must be yes/no/unclear with EXACT answer format:
  "Answer exactly: 'yes', 'no', or 'unclear'."
- Include at least one generic count question with the fixed answer set:
  "Answer exactly: 'one', 'two', 'many', 'none', or 'unclear'."
- You may add 1--3 short fill-in questions (appearance, color, scene) with answer formats like:
  "Answer in up to two words."
  "Answer in at most five words."
- Do NOT introduce new specific objects that do not appear in the short caption or detailed caption.

Probe for contradictions:
- If the short caption claims something unusual or very specific (e.g., "two women", "three giraffes", "a bird pulling a wagon", "an old telephone in a bathtub"), design questions that directly test those claims, especially the number of subjects and the main object's appearance.
- For relations (e.g., "holding", "sitting on", "next to", "leaning against"), ask explicitly whether this relation is true.

Output format (IMPORTANT):
- Return ONLY a single JSON object:
  {
    "sub_questions": [ "...", ... ]
  }
- "sub_questions" must be a list of 3-5 question strings.
- Each question string must include its own answer-format instruction.
- Do NOT add any explanation or extra keys.
\end{promptbox}

For caption-based evaluation, the decomposition prompt is specialized to verify the structured claims produced in the previous step: given the extracted subject, count, key object, relation, and scene, it asks the LLM to generate 3-5 focused verification questions with fixed answer formats, making it easy to detect contradictions between the short caption and the image while keeping the output as a clean JSON list of queries for the VLM.

\subsection{Prompts for Aggregation}
\label{subsec:prompt_aggregation}

We finally specify the prompts that aggregate sub-question answers into a single prediction. For generic VQA-style tasks, the aggregator reasons over all sub-answers to produce a short, task-aligned answer; for captioning, the aggregator acts as a COCO-style caption judge that decides whether to keep or minimally correct the original caption.

\noindent\textbf{(a) System prompt (Aggregation for VQA/ImageNet-D).}

\begin{promptbox}
Answer the question by reasoning over ALL sub-question answers. Follow these rules:

1. Output format:
   - Quantity questions: ONLY the number (e.g., "2")
   - Yes/no questions: ONLY "yes" or "no"
   - Action questions ("what is doing"): ONLY the verb/phrase (e.g., "crossing" not "crossing the street")
   - Gaze questions ("where looking"): ONLY direction (up/down/left/right, not "-wards" or "upwards")
   - NO articles: a/an/the
   - Others: single word/phrase (max 3--4 words)

2. Reasoning process (internal):
   a. Synthesize ALL answers to form a coherent conclusion
   b. Make logical inferences (e.g., trees + leaves -> forest)
   c. Avoid being too dependent on any single sub-question

3. For location/context questions:
   - Combine environmental clues (e.g., trees + leaves -> forest)
   - Consider object interactions (e.g., beach + waves -> ocean)

Question:
{question}

Sub-questions and answers:
{sub}

Output ONLY the answer with no additional text (no explanations):
\end{promptbox}

For VQA and ImageNet-D, the aggregation prompt instructs the LLM to treat sub-question answers as evidence, integrate them with lightweight internal reasoning, and return a minimal, format-controlled answer string (number, yes/no, or short phrase), which keeps the output easy to score while reducing sensitivity to any single noisy or adversarial probe.

\noindent\textbf{(b) System prompt (Aggregation for Caption Verification).}

\begin{promptbox}
COCO-style caption judge and editor (no trimming + confidence gate).

Inputs:
1. A short COCO-style caption (one sentence).
2. Parsed claims from that caption: subject_head, subject_count, key_object, relation, scene.
3. A small set of sub-questions and their answers from a VLM that can see the image.

Critical slots to check:
- main subject identity (head noun)
- subject count (one/two/many)
- key object (if any)
- relation between subject and key object
- coarse scene type (if the caption asserts one)

Strict decision rule with confidence gate:

For each slot that the short caption clearly asserts:
- Look for support in the answers.
  - "yes" on a matching question = positive support.
  - Any "no" that contradicts the slot = conflict.
  - Only "unclear" (and no supporting "yes") = unsupported.

Confidence gate (edit only when clearly confident):
You may modify the caption only if at least one is true:
1. Direct conflict: a "no" explicitly contradicts the asserted slot.
2. Strong counter-evidence for a different value: at least two consistent, relevant "yes" answers indicating a specific alternative, with no conflicting "no" answers for that slot.

If a slot is merely unsupported (only "unclear" or weak/mixed signals) and not contradicted, do not change that slot.

If all asserted slots are non-conflicting and at least one asserted slot has positive support, keep the original caption.

Global decision:
- Set has_conflict = true only when you actually change the caption under the confidence gate above.
- Otherwise set has_conflict = false and keep the caption exactly as given.

Edit rules (no trimming policy):

When has_conflict = true:
- Preserve all original descriptive content whenever possible.
  - Only change words that are explicitly proven wrong by the sub-question answers.
  - Do not shorten, simplify, or remove adjectives, modifiers, or secondary objects unless they are directly contradicted.
  - If uncertain, keep the original content instead of deleting it.

Allowed edits:
1. Fix only the specific critical slot(s) that conflict:
   - subject_head (e.g., "dog" -> "cat")
   - subject_count (e.g., "two" -> "one")
   - key_object (e.g., "table" -> "bench")
   - relation (e.g., "on" -> "next to")
   - scene (e.g., "kitchen" -> "living room")
2. Keep all other content exactly unchanged.
3. You may rephrase minimally to maintain grammatical correctness after fixing the conflicts.
4. Do not trim sentence length, remove descriptors, or generalize nouns unless you are absolutely certain that the original word is wrong.

Style:
- One English sentence, 8--20 words (slightly longer allowed for full retention).
- Preserve original detail level and structure whenever possible.
- Use natural, descriptive COCO-style language.

Output format (IMPORTANT):
Return only a JSON object with exactly these keys:
{
  "has_conflict": <true or false>,
  "caption": "<final short caption>"
}

Important:
- If "has_conflict" is false, "caption" must be exactly the original short caption (no rephrasing, no added words, no removed words).
- If "has_conflict" is true, "caption" must be the corrected sentence following the rules above, changing only the slots that are clearly contradicted while preserving all other descriptive content.
\end{promptbox}

For caption-based aggregation, the prompt casts the LLM as a conservative COCO-style caption judge that uses VLM-derived answers to check a small set of critical slots and only edits the caption when there is clear, slot-specific evidence of conflict, otherwise returning the original sentence verbatim; the JSON output with a boolean conflict flag and a single caption string makes it straightforward to integrate this decision into our overall PDA pipeline.

\begin{figure}[t]
    \centering
    \includegraphics[width=\linewidth]{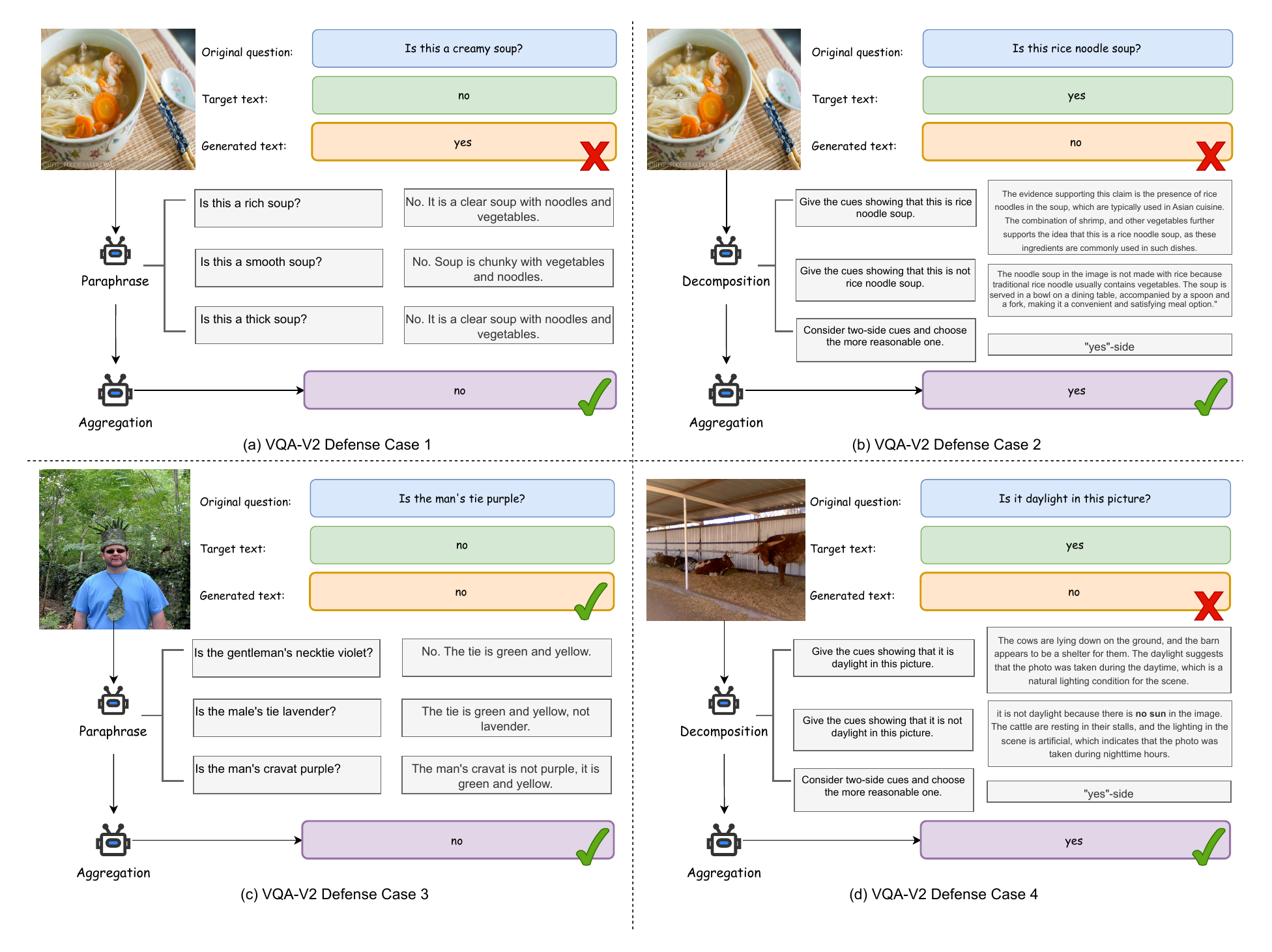}
    \caption{Additional VQA-v2 qualitative examples. For each adversarial image we compare the answer of the undefended VLM with the output of PDA and visualize key paraphrases and sub-questions that support the corrected decision.}
    \label{fig:supp_vqa_examples}
\end{figure}

\begin{figure}[t]
    \centering
    \includegraphics[width=\linewidth]{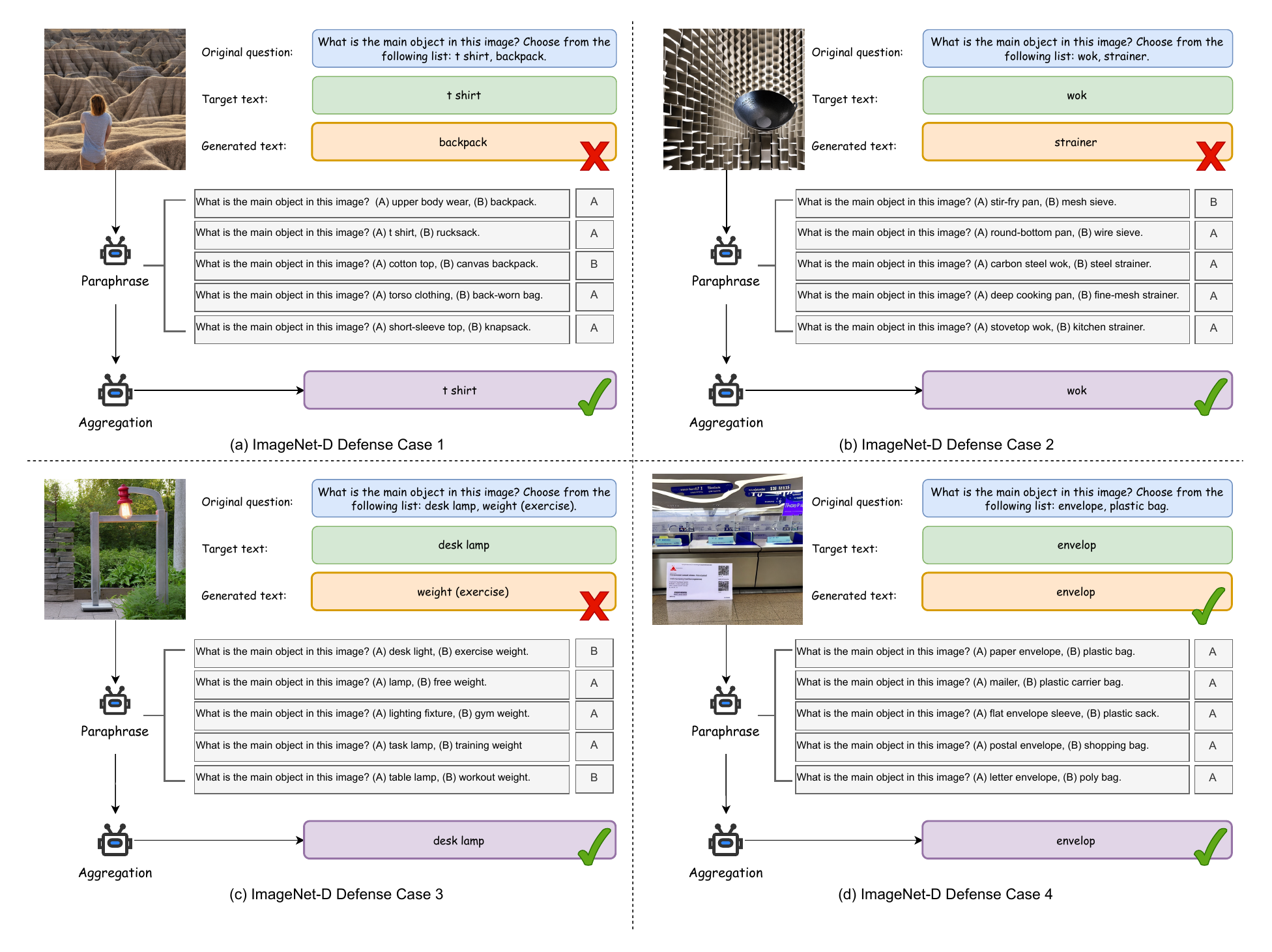}
    \caption{Additional ImageNet-D qualitative examples. PDA stabilizes fine-grained recognition by aggregating evidence from multiple paraphrases that describe shape, part configuration, and material, instead of relying on a single corrupted view.}
    \label{fig:supp_imagenetd_examples}
\end{figure}

\begin{figure}[t]
    \centering
    \includegraphics[width=\linewidth]{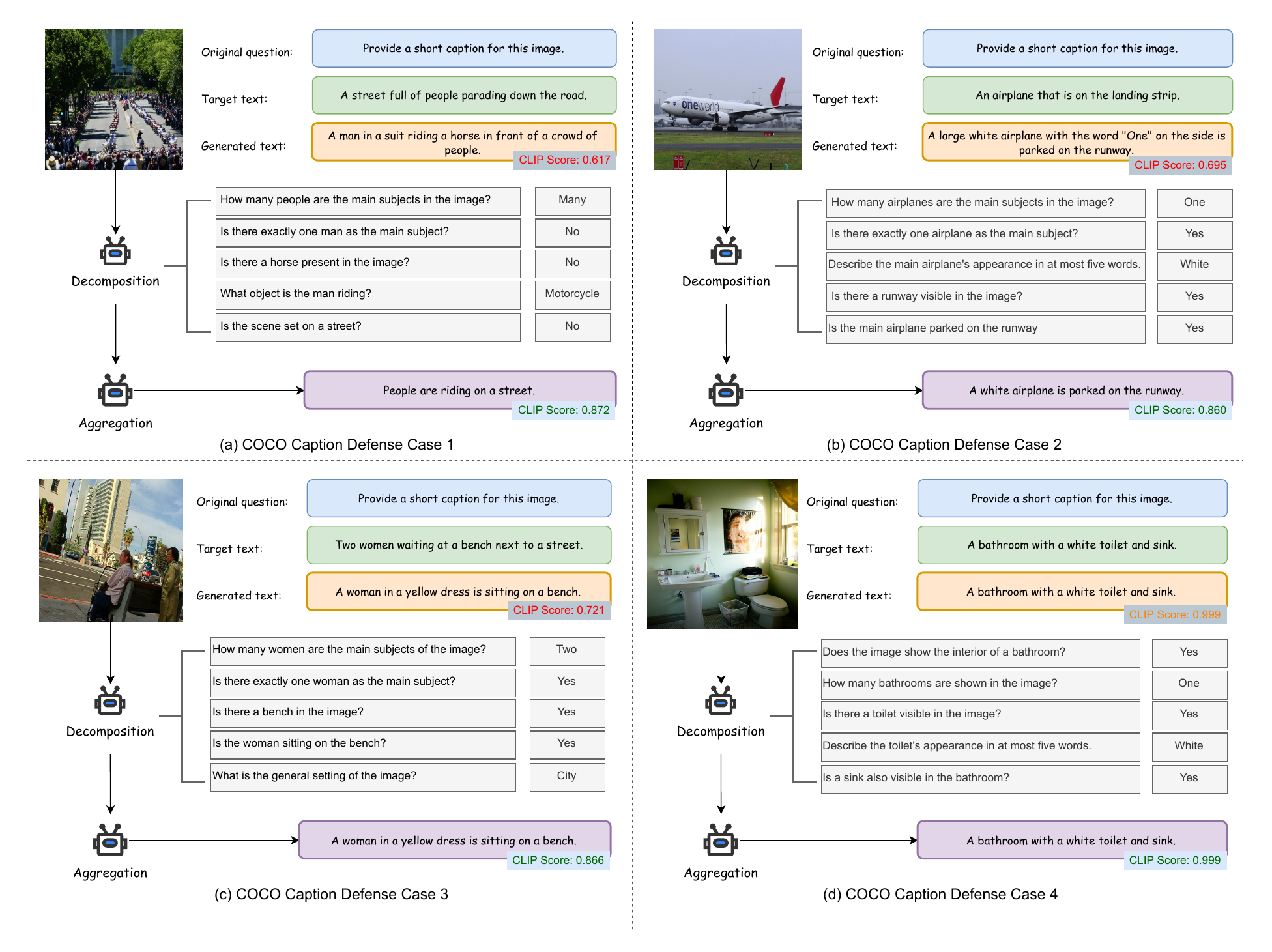}
    \caption{Additional COCO captioning examples. PDA detects and repairs conflicts between a corrupted short caption and VLM answers to verification sub-questions, yielding minimally edited but visually consistent captions.}
    \label{fig:supp_coco_examples}
\end{figure}

\section{Additional Qualitative Examples}
\label{sec:supp_qualitative}
This section provides additional qualitative examples across VQA-v2, ImageNet-D, and COCO captioning to complement the quantitative results in the main paper. For each dataset we highlight typical failure modes exhibited by undefended VLMs and how PDA corrects them through paraphrasing, decomposition, and aggregation. All examples are drawn directly from our evaluation pipeline.

\subsection{VQA-v2: Defense Case Studies}
\label{subsec:supp_vqa}
Figure~\ref{fig:supp_vqa_examples} shows several representative VQA-v2 cases where adversarial perturbations cause the victim models to produce incorrect answers on seemingly easy questions, such as misclassifying food categories or confusing simple attributes (e.g., soup type, clothing color, or whether the scene is in daylight). PDA corrects these failures by enforcing semantic consistency across multiple paraphrases and their associated sub-questions: even when some prompts are hijacked by the attack, the majority of paraphrase--sub-question pairs still support the correct answer, and the aggregation stage downweights inconsistent evidence and recovers the underlying semantics.

\subsection{ImageNet-D: Fine-Grained Recognition Under Shift}
\label{subsec:supp_imagenetd}
On ImageNet-D, the combination of distribution shift and adversarial noise often pushes the baseline models toward visually nearby but incorrect categories. As illustrated in Figure~\ref{fig:supp_imagenetd_examples}, undefended VLMs tend to overfit to corrupted local textures or background cues, leading to fine-grained misclassification. PDA mitigates this by forcing agreement between paraphrase-derived cues that describe shape, parts, material, and coarse category; inconsistent labels that cannot be reconciled with these multi-view descriptions are rejected, and the final aggregated prediction aligns more closely with the true object class, even when the raw logits of the victim model are heavily distorted.

\subsection{COCO Captioning: Caption Repair and Conflict Resolution}
\label{subsec:supp_coco}
For COCO captioning, the task is to repair a short caption that has been corrupted by adversarial perturbations. Figure~\ref{fig:supp_coco_examples} presents several examples where the baseline caption either hallucinates non-existent objects, misstates the subject identity or count, or confuses relations and scene type. PDA first extracts atomic claims from the short caption (main subject, subject count, key object, relation, and scene) and then verifies each claim using a set of structured sub-questions posed to a VLM that can see the image. Claims that conflict with the answers are minimally edited, while unsupported but non-contradicted content is left unchanged. As a result, PDA produces corrected captions that fix concrete visual errors without aggressively rewriting or shortening the original sentence, leading to more reliable and faithful descriptions under adversarial conditions.